\pdfoutput=1
\documentclass[10pt,twocolumn,letterpaper]{article}

\usepackage[pagenumbers]{cvpr}

\usepackage{cuted}        
\usepackage[T1]{fontenc}
\usepackage[scaled=0.9]{helvet} 
\usepackage{microtype}
\usepackage{multirow}
\usepackage{makecell}
\usepackage{bm}
\usepackage{tabularx}

\renewcommand{\paragraph}[1]{\vspace{2pt}\noindent\textbf{#1}\hspace{0.45em}}

\newcommand{\best}[1]{\textbf{#1}}
\newcommand{\second}[1]{\underline{#1}}   
\newcommand{\na}{--}
\definecolor{ceilinggray}{gray}{0.42}     

\AtEndPreamble{\AtBeginDocument{%
  \crefname{appendix}{App.}{Apps.}%
  \Crefname{appendix}{Appendix}{Appendices}}}

\newcommand{\InsideJEPA}{Inside-JEPA\xspace}

\newcommand{\ArmFormat}[1]{\textsf{#1}}
\newcommand{\PRED}{\ArmFormat{PRED}\xspace}          
\newcommand{\INVS}{\ArmFormat{INV-S}\xspace}         
\newcommand{\INVZ}{\ArmFormat{INV-Z}\xspace}         
\newcommand{\INVR}{\ArmFormat{INV-R}\xspace}         
\newcommand{\INVSHUF}{\ArmFormat{INV-SHUF}\xspace}   
\newcommand{\AOT}{\ArmFormat{AOT-S}\xspace}          
\newcommand{\NCE}{\ArmFormat{NCE-S}\xspace}          
\newcommand{\SUP}{\ArmFormat{SUP-S}\xspace}          

\newcommand{\Ljepa}{\mathcal{L}_{\mathrm{JEPA}}}
\newcommand{\Linv}{\mathcal{L}_{\mathrm{inv}}}
\newcommand{\Ltot}{\mathcal{L}}
\newcommand{\sg}{\operatorname{sg}}
\newcommand{\Ebar}{\bar{E}}

\newcommand{\AUROC}{\operatorname{AUROC}}
\newcommand{\mAUROC}{\operatorname{mAUROC}}
\DeclareMathOperator*{\argmax}{arg\,max}

\newcommand{\Rset}{\mathcal{R}}   
\newcommand{\Aset}{\mathcal{A}}   
\newcommand{\R}{\mathbb{R}}

\newcommand{\rel}[1]{\textsf{\detokenize{#1}}}          
\newcommand{\PlDist}{\textsf{Dist}\xspace}
\newcommand{\PlChan}{\textsf{Chan}\xspace}
\newcommand{\PlInv}{\textsf{Inv}\xspace}
\newcommand{\PlZ}{\textsf{Probe-Z}\xspace}
\newcommand{\PlS}{\textsf{Probe-S}\xspace}
\newcommand{\PlR}{\textsf{Probe-R}\xspace}
\newcommand{\Crule}[1]{\textsf{C#1}\xspace}               

\newcommand{\up}{\ensuremath{\uparrow}}

\newcommand{\twoSE}{2\,SE\xspace}

\definecolor{cvprblue}{rgb}{0.21,0.49,0.74}
\usepackage[pagebackref,breaklinks,colorlinks,allcolors=cvprblue]{hyperref}
\hypersetup{pdftitle={Inside-JEPA: What Makes Relational Channels Meaningful? Inverse Dynamics, Not Latent Prediction}}

\title{\InsideJEPA: From Prediction to Relational Grounding \\ in Latent World Models}

\author{
Hossein Resani$^{1}$ \hspace{3em}
Javen Qinfeng Shi$^{1,2}$\\
$^{1}$Australian Institute for Machine Learning, Adelaide University\\
$^{2}$Responsible AI Research Centre
}

\begin{document}
\maketitle

\begin{strip}
\centering
\includegraphics[width=\textwidth]{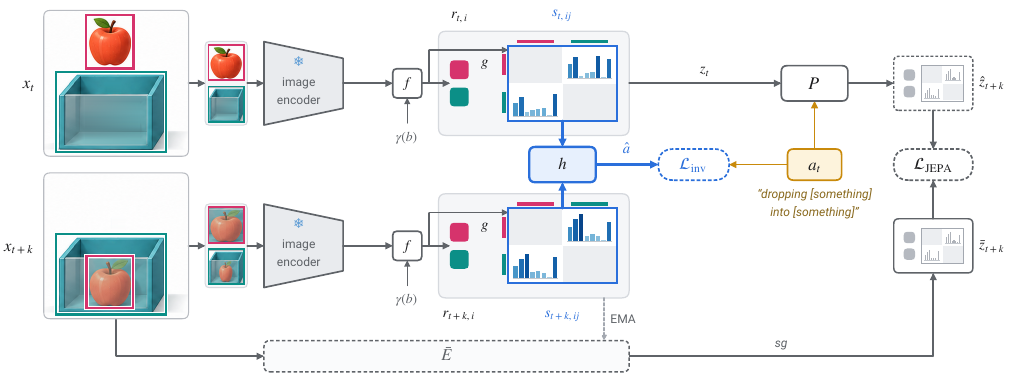}
\captionsetup{hypcap=false}
\captionof{figure}{\textbf{\InsideJEPA.} Each entity box $b_{t,i}$ is cropped from frame $x_t$, and $f$ combines the crop's features from a frozen image encoder with a Fourier embedding $\gamma(b)$ of the box. Its output gives the entity latent $r_{t,i}$ and, through a relation head $g$ on every ordered pair, $K$ relational channels $s_{t,ij}\in(0,1)^K$ (grid rows and columns: entities $i$ and $j$). The state $z_t=(r_t,s_t)$ is advanced by a predictor $P$ and trained, as in every model we compare, to match an exponential-moving-average target encoder $\Ebar$ under stop-gradient ($\Ljepa$, grey). \InsideJEPA adds an inverse-dynamics head $h$ that must recover the action $a_t$, here the Something-Something template ``dropping [something] into [something]'', from the relational channels $s_t$ and $s_{t+k}$ (blue), so the action labels decide what the channels come to mean. No relation label is used in training.}
\captionsetup{hypcap=true}
\label{fig:overview}
\end{strip}

\begin{abstract}
Joint-embedding predictive architectures (JEPAs) learn world models in latent space, enabling efficient planning but leaving their states opaque.
We introduce \InsideJEPA which places relational channels for ordered entity pairs inside the JEPA latent state and show that an inverse-dynamics loss can ground these channels without relation labels when its head reads them, either alone or as part of the full state.
On Something-Something V2, latent prediction alone performs no better than a random relation head, while inverse-dynamics supervision that recovers actions from channel changes improves relational AUROC by .065--.103 across ten settings with annotated boxes, by .081 on RH20T robot episodes with held-out tasks, and by .050--.061 across five relations using off-the-shelf detection and tracking.
An inverse head on entity latents decodes actions more accurately but barely improves relational grounding; shuffled labels, two label-free objectives, and action labels on only 5\% of the pairs also fail to ground the channels.
Given an action, predicted relational changes approach the quality of observed future relations (AUROC .599 \vs\ .621).
For one-step relational goals, a supervised readout of the 16 channels ranks the correct action among the top five of 65 in 43\% of cases, compared with 24\% for prediction-only latents and 17\% for a frequency prior.
\end{abstract}

\section{Introduction}
\label{sec:intro}

World models that predict in a learned latent space rather than in pixels can ignore what is unpredictable and keep what matters for acting~\cite{lecun2022path,assran2023ijepa,bardes2024vjepa}, and they now support planning from images and video~\cite{assran2025vjepa2,zhou2025dinowm,bar2025nwm}.
Their states, however, are opaque vectors.
A probe trained on relation labels can often read from them whether the cup is inside the bowl, but no part of the state is committed to that relation, so neither a goal nor what the model believes can be stated in those terms, although relations between objects are the natural vocabulary for both~\cite{xu2017scenegraph,ji2020actiongenome,yi2020clevrer}. 
In our real-video evaluations, these relations are geometric predicates computed from 2D entity bounding boxes, so \textit{inside} denotes an image-plane proxy for containment.

A direct remedy is to build the relations into the state: for every ordered pair of entities, a small vector of soft symbols, which we call relational channels, that the predictor consumes and forecasts like the rest of the latent.
Unlike neuro-symbolic pipelines that read predicates out of a representation with supervised heads~\cite{koh2020cbm,xu2017scenegraph,mao2019nscl}, such channels sit inside the dynamics and are trained only through the world model's own objectives, with no relation labels.
Nothing in a latent-prediction objective, however, says what they should mean.
A predictor is as well served by a static code, or by one that changes without tracking anything, as by a code for the relations: the symbol grounding problem~\cite{harnad1990symbol}, inside a world model.
To our knowledge, no prior work has tested whether relational channels in a world model learn meaningful relationships from real videos, or which training signals help them do so.

\begin{figure*}[t]
\centering
\includegraphics[width=\linewidth]{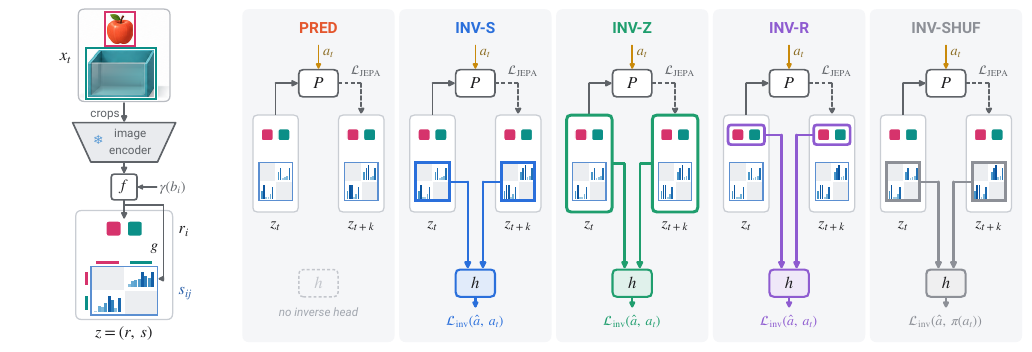}
\caption{\textbf{Five arms, one difference.} All arms share encoder, predictor, data, budget and loss weight and differ only in what the inverse head $h$ reads, and so in where its gradient enters the state (coloured outlines), or in its targets. \PRED has no inverse head. \INVS (\InsideJEPA) reads the relational channels $s$; \INVZ reads the whole latent $z=(r,s)$; \INVR reads only the entity latents $r$, so its gradient reaches $s$ only through the shared encoder; \INVSHUF is \INVS with shuffled targets $\pi(a_t)$ for the inverse loss only.
$P$ receives $a_t$ when action conditioning is enabled. The action inputs shown are omitted in the first four settings of \cref{tab:annot} and in the action-free forecasting control.}
\label{fig:arms}
\end{figure*}

We study this question with \InsideJEPA(\cref{fig:overview}).
Our method adds one term to the JEPA objective, an inverse-dynamics loss in which a head must recover the action from the relational channels before and after it, so that the action labels an action-conditioned world model consumes anyway must explain how its relational channels change.
To find out what grounds the channels, we compare five models that differ only in whether and where the inverse gradient enters the state and in whether its targets are shuffled (\cref{fig:arms}), plus label-free and label-fraction controls.
Because post-hoc channel matching makes it easy to find meaning that is not there, each relation is matched to a single channel, and every model is also scored with its relation head replaced by a randomly initialised one, which sees the same box geometry (\cref{fig:protocol}).

Across a controlled 2-D world, Something-Something V2 with Something-Else boxes~\cite{goyal2017ssv2,materzynska2020somethingelse}, RH20T robot episodes~\cite{fang2023rh20t} and objects found by an off-the-shelf detector and tracker~\cite{liu2024groundingdino,ravi2025sam2}, the pattern is the same (\cref{fig:grounding}).
Latent prediction alone leaves the channels at or near the level of a random relation head, or below it.
Inverse dynamics grounds them: matched relational AUROC rises by .065--.103 in ten settings with annotated boxes, by .081 on robot episodes with end-effector actions and held-out tasks, and by .050--.061 with tracked objects on five relations; the gain is positive in all 18 settings of \cref{fig:grounding} and smallest with tracked objects on thirteen relations and with a frozen V-JEPA~2 encoder (+.016 to +.042).
What grounds them is correct action labels that the channels themselves must explain: an inverse head on the entity latents decodes the action best of all models yet barely moves the channels, and shuffled labels, two label-free objectives and labels on 5\% of the pairs do not ground them.
The grounded channels are useful: given an action, their forecasts of the relations that change score almost as well as the observed future, and a readout of the 16 channels selects actions towards relational goals better than a readout of the prediction-only model's whole latent (top-5 .425 \vs\ .244, frequency prior .174).

\noindent Our contributions are:
\begin{itemize}[leftmargin=*,itemsep=1pt,topsep=2pt]

\item \textbf{\InsideJEPA}, a JEPA world model with relational channels inside its latent state, and evidence that an inverse-dynamics loss can ground those channels without relation labels when its head reads them, either alone (\INVS) or as part of the full state (\INVZ)  (\cref{sec:method}).

\item \textbf{A controlled account of what grounds the channels}: five models that differ only in their inverse head, on two datasets with two kinds of action and with automatically found objects, scored by one-to-one matched AUROC and against a random relation head. Grounding needs an inverse head that reads the channels and correct action labels on thousands of pairs; latent prediction, shuffled labels and two label-free losses do not ground them (\cref{sec:rq1,sec:rq2}).
\item \textbf{Uses and limits of the grounded channels}: action-conditioned forecasting and one-step action selection towards relational goals on Something-Else, and a measured account of where grounding stops: left/right, a distributed code, a stronger encoder and a supervised ceiling (\cref{sec:rq3,sec:rq4}).
\end{itemize}

\section{Related work}
\label{sec:related}

\paragraph{Latent-predictive world models.}
JEPAs predict target-encoder embeddings instead of pixels~\cite{lecun2022path,assran2023ijepa,bardes2024vjepa}; V-JEPA~2 adds action conditioning and plans with it~\cite{assran2025vjepa2}; DINO-WM and navigation world models plan in the feature space of pretrained encoders~\cite{zhou2025dinowm,bar2025nwm};
and model-based RL learns action-conditioned latent dynamics~\cite{hafner2023dreamerv3,ha2018worldmodels,hansen2024tdmpc2}.
All keep unstructured vector states.
We keep the JEPA objective and ask what relational channels in its state come to mean.

\paragraph{Objects and relations in dynamics models.}
Interaction networks model physics over given objects and relations~\cite{battaglia2016interaction}; C-SWM learns object-factored transitions with a contrastive loss~\cite{kipf2020cswm}; slot models discover objects without boxes~\cite{locatello2020slot,kipf2022savi,seitzer2023dinosaur} and roll them forward~\cite{wu2023slotformer}.
Scene graphs and their spatio-temporal versions describe pairwise relations with supervised predicates~\cite{xu2017scenegraph,ji2020actiongenome}, and neuro-symbolic reasoners execute programs over such structure~\cite{mao2019nscl,yi2020clevrer}.
Our relational channels are pairwise like a scene graph but trained without relation labels and part of the dynamics state; our learned-object experiments try slots, then a detector and tracker.

\paragraph{Symbols for planning.}
Latplan learns propositional states from images for classical planning~\cite{asai2018latplan}, and skill-to-symbol methods derive symbols from the preconditions and effects of actions~\cite{konidaris2018skills}.
Closest to our setting, Migimatsu and Bohg ground predicate classifiers on Something-Something~V2 with Something-Else boxes from action labels alone, turning each action's hand-written PDDL pre- and post-conditions into partial labels~\cite{migimatsu2022grounding}.
Our inverse loss also ties the channels to what actions do, but needs no hand-written conditions or separate classifier: it is a differentiable term in a JEPA, and its channels are tested on relation labels it never sees.
Because that method needs hand-written conditions for every action, we do not run it; our comparisons hold the architecture fixed and vary only the training signal.

\paragraph{Inverse dynamics as a learning signal.}
Predicting the action from consecutive observations shapes features for manipulation and exploration~\cite{agrawal2016poke,pathak2017icm}, pretrains representations for imitation~\cite{brandfonbrener2023idm} and labels or infers actions in unlabelled video~\cite{baker2022vpt,schmidt2024lapo}.
Recent work has also incorporated inverse dynamics into JEPA-style world models: as the sole anti-collapse regulariser~\cite{ivashkov2026smwm}, as a contrastive head replacing SIGReg~\cite{boylan2026acmtm}, as a decoder of actions from latent differences~\cite{zhang2026deltajepa}, or to keep them plannable when little of the frame moves~\cite{strohm2026plannable}; for linear systems, exact action reconstruction provably keeps every state direction reachable within the horizon~\cite{toso2026unstable}.
They shape the whole latent for control, as our \INVZ arm does; we use the term to give chosen channels of the state a meaning, and measure whether it does.

\paragraph{Concepts, probes and identifiability.}
Concept-bottleneck models supervise concept channels directly~\cite{koh2020cbm}; linear probes test what a representation encodes~\cite{alain2016probes}; and unsupervised factors are not identifiable without inductive bias or supervision~\cite{locatello2019challenging}.
We use linear probes, never in training, to check that relational information is present and as planning readouts, and a relation-supervised version of our channels only as a ceiling.
Temporal self-supervision, arrow-of-time classification~\cite{misra2016shuffle,wei2018arrow} and contrastive prediction~\cite{oord2018cpc}, provides our label-free controls; our InfoNCE control contrasts start and goal channels without the action.

\section{\texorpdfstring{\InsideJEPA}{Inside-JEPA}}
\label{sec:method}

\InsideJEPA is a JEPA world model whose state contains one latent per entity and one vector of relational channels per ordered entity pair (\cref{fig:overview}).
It is trained with the JEPA prediction loss plus an inverse-dynamics loss on the relational channels; no relation label is used.

\subsection{A relational latent state}
\label{sec:state}
A training example is a pair of frames $(x_t, x_{t+k})$ from one clip, the boxes $b_{t,i}$ of the $N\le 4$ entities (objects and hands) present in both frames, and the action $a_t$ taken in between.
Each entity is encoded as
\begin{equation}
  e_{t,i} = f\big(\phi(x_t[b_{t,i}]),\, \gamma(b_{t,i})\big), \quad r_{t,i} = \mathrm{LN}(W e_{t,i}),
  \label{eq:entity}
\end{equation}
where $x_t[b]$ is box $b$ cropped from $x_t$ and resized to \mbox{$96\times96$}, $\phi$ is a frozen ImageNet ResNet-18~\cite{he2016resnet}, $\gamma$ is a Fourier-feature embedding of the box coordinates~\cite{tancik2020fourier}, $f$ and $W$ are learned, LN is LayerNorm, and $r_{t,i}\in\R^{32}$.
A relation head $g$ maps every ordered pair of distinct entities to $K$ relational channels, soft symbols in $(0,1)$,
\begin{equation}
  s_{t,ij} = \sigma\big(g(e_{t,i}, e_{t,j})\big) \in (0,1)^{K}, \qquad i \neq j,
  \label{eq:symbols}
\end{equation}
with $K=8$ in experiments with five relations and $K=16$ with thirteen.
The state $z_t=(r_t,s_t)$ concatenates all entity latents and relational channels; entries of padded entities and of $i=j$ are held constant.
Nothing ties a channel of $s$ to a relation, and the channels are part of the state the predictor consumes and forecasts.
Because box coordinates enter $e$, even an untrained $g$ sees geometry, so every model is also scored with its relation head replaced by a randomly initialised one (\cref{sec:protocol}).

\subsection{Latent prediction}
\label{sec:jepa}
A target encoder $\Ebar$, an exponential moving average of the online encoder with momentum 0.996~\cite{grill2020byol,assran2023ijepa}, embeds the later frame as $\bar z_{t+k}$, and a predictor $P$ forecasts it:
\begin{equation}
  \Ljepa = \big\| P\big(z_t, u(a_t)\big) - \sg(\bar z_{t+k}) \big\|_2^2 ,
  \label{eq:jepa}
\end{equation}
where $\sg$ is stop-gradient, $u$ a learned 64-d action embedding (a linear map of the control vector on robot data) and $P$ a two-layer MLP on the flattened state.
$P$ receives the action in every experiment, as drawn in \cref{fig:overview}, except the first four settings of \cref{tab:annot} and the test of forecasting without the action.
\cref{eq:jepa} alone defines the prediction-only model \PRED.
It constrains the channels only to make $\bar z_{t+k}$ predictable, which a static code, or one that changes without tracking the scene, satisfies as well as a relational one.

\subsection{Inverse dynamics on the relational channels}
\label{sec:inverse}
\InsideJEPA adds an inverse head $h$ that must recover the action from the relational channels before and after it:
\begin{equation}
  \Linv = \ell\big(h(s_t, s_{t+k}),\ a_t\big), \qquad \Ltot = \Ljepa + \lambda\,\Linv ,
  \label{eq:inv}
\end{equation}
where $\ell$ is the cross-entropy over action classes (40, 65 or 174 Something-Something templates) or the squared error for continuous robot controls, and $\lambda=1$ in every experiment.
$s_{t+k}$ comes from the online encoder, so the gradient of $\Linv$ reaches $g$ and $f$ at both frames; $\Ebar$ provides only the target of \cref{eq:jepa}.

\paragraph{Design rationale (a hypothesis that \cref{sec:experiments} tests).}
The change of a static or arbitrary code carries no information about the action, so the codes that satisfy \cref{eq:jepa} trivially cannot minimise \cref{eq:inv}.
Relations such as contact and containment change in ways that identify manipulation actions, so a code that tracks them helps recover the action.
\cref{eq:inv} does not say which channel should carry which relation, nor anything about relations whose change does not reveal the action.

\subsection{Arms and controls}
\label{sec:arms}

To identify what grounds the channels, we compare five arms (\cref{fig:arms}) that share architecture, data, training budget and $\lambda$, with no per-arm tuning.
\PRED uses \cref{eq:jepa} only.
\INVS is \InsideJEPA.
\INVZ lets $h$ read the whole latent $(z_t, z_{t+k})$, as inverse-dynamics terms in recent JEPA world models do (\cref{sec:related}), so its gradient also reaches the channels.
\INVR lets $h$ read only the entity latents $(r_t, r_{t+k})$, so its gradient reaches the channels only through the shared encoder $f$.
\INVSHUF is \INVS with shuffled action targets.
Three further controls act on the same channels: \AOT and \NCE replace $\Linv$ with label-free losses, an arrow-of-time classifier~\cite{misra2016shuffle,wei2018arrow} and InfoNCE between start and goal channels~\cite{oord2018cpc}, each weighted to match $\Linv$ at initialisation; \SUP trains channel $\rho$ of $s$ directly on relation $\rho$ and serves only as a ceiling.

\section{Evaluation protocol}
\label{sec:protocol}

\begin{figure*}[t]
\centering
\includegraphics[width=\linewidth]{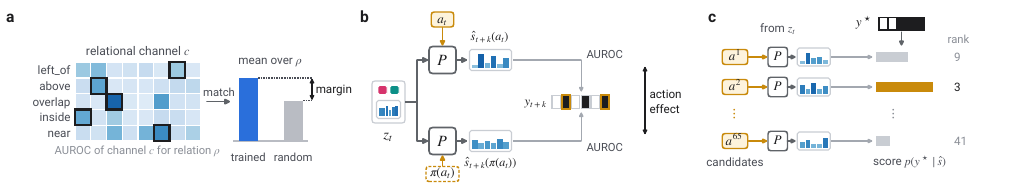}
\caption{\textbf{How we test the relational channels} (schematic). \textbf{(a)} Grounding: the AUROC of every relational channel $c$ for every relation $\rho$ is computed on validation pairs; Hungarian matching gives each relation one channel, and the mean matched AUROC is compared with that of a randomly initialised relation head (margin). \textbf{(b)} Forecasting relational change: from $z_t$, $P$ forecasts $\hat s_{t+k}$ with the true action $a_t$ or a permuted one $\pi(a_t)$; both are scored on the entries of $y_{t+k}$ whose relation changes (outlined), and their difference is the action effect. \textbf{(c)} Planning: every candidate action is rolled through $P$, each forecast is scored by the likelihood of the relational goal $y^\star$ under a readout, and we report how often the true action ranks in the top five.}
\label{fig:protocol}
\end{figure*}

\paragraph{Relations.}
Binary relations between entity boxes are computed on every frame and used only for evaluation: five (\rel{left_of}, \rel{above}, \rel{overlap}, \rel{inside}, \rel{near}) or thirteen, which add \rel{far}, \rel{near_tight}, \rel{y_overlap}, \rel{a_covers_half_b}, \rel{much_larger}, \rel{strictly_above} and two hand relations, \rel{hand_near_obj} and \rel{hand_overlaps_obj}.
Let $y^{\rho}_{t,ij}\in\{0,1\}$ denote relation $\rho\in\Rset$ for the ordered pair $(i,j)$.

\paragraph{Matched relational AUROC} (\cref{fig:protocol}a).
For relation $\rho$ and relational channel $c$, let $A_{\rho c}$ be the AUROC of $s^{c}_{ij}$ for $y^{\rho}_{ij}$ over all valid ordered pairs of the validation frames, with no sign flip.
Hungarian matching selects an injective map $m^\star=\argmax_{m:\Rset\to\{1..K\}}\sum_{\rho}A_{\rho,m(\rho)}$, and
\begin{equation}
  \mAUROC = \frac{1}{|\Rset|}\sum_{\rho\in\Rset} A_{\rho,\,m^\star(\rho)} .
  \label{eq:mauroc}
\end{equation}
This is our headline metric.
It is strict: one channel per relation, against relations computed from boxes; it therefore understates a relation spread over several channels (\cref{sec:rq4}).
The matching is fitted on the validation pairs it scores; fitting it on training pairs instead leaves planning with the matched channels unchanged (within .015).
The \emph{grounding gain} is \INVS's $\mAUROC$ minus \PRED's.
Because geometry enters every state, we also score each trained model with its relation head $g$ replaced by a randomly initialised one (seed 0) and report the \emph{margin}, trained minus random-head $\mAUROC$.
A linear probe on each arm's pair features checks that the relations are decodable at all.
We call the channels \emph{grounded} when they align with the box-computed relations beyond both the prediction-only model and a random relation head, so besides the gain we report the \emph{margin gain}, an arm's margin minus \PRED's.
Unless stated otherwise, results are means over three seeds; SE is the standard error over seeds, and \twoSE, twice it, is a lenient interval with three to five seeds.

\paragraph{Forecasting} (\cref{fig:protocol}b).
From $z_t$ and an action, $P$ forecasts $\hat z_{t+k}$, and the matched channels of $\hat s_{t+k}$ are scored by AUROC against $y_{t+k}$ on the entries whose relation changes between $t$ and $t+k$, so the score measures how well the forecast gets the new values of changing relations right.
References are copying $s_t$, the channels observed at $t+k$ and a forecast with a permuted action; the \emph{action effect} is the AUROC with the true action minus that with the permuted one.

\paragraph{Planning} (\cref{fig:protocol}c).
Planning is one-step action selection: given $z_t$ and a goal $y^\star=y_{t+k}$, the clip's own end-state relations over all pairs, every candidate $a\in\Aset$ is ranked either by the likelihood of $y^\star$ under a per-relation readout of its forecast from $P$ or by the arm's inverse head.
\PlChan reads each relation's matched channel through a logistic calibration; \PlInv feeds $s_t$ and the goal, mapped into channel space, to the arm's own inverse head; \PlZ, \PlS and \PlR are linear logistic probes on the forecast pair features $(r_i, r_j, s_{ij})$, on $s_{ij}$ only, or on $(r_i, r_j)$ only.
Readouts are fitted on relation-labelled training pairs, so planning, unlike grounding, uses relation labels.
We report the top-5 accuracy of the true action among the 65 templates, against chance (.077) and a template-frequency prior (.174).
We use calibrated readouts because an evaluation-only analysis of earlier 40-template checkpoints showed that raw squared-distance scoring hid the planning signal (\cref{sec:evaldetails}).

\paragraph{Data.}
\emph{Something-Something V2} (SSv2)~\cite{goyal2017ssv2} \emph{with Something-Else boxes}~\cite{materzynska2020somethingelse}, Something-Else for short, compositional split: the 40 most frequent relational templates (8,000/2,000 videos), all 65 relational templates (13,954 training pairs, every annotated training video for them; 14,975 validation pairs) or, in one replication, all 174 templates (42,164/44,342 pairs).
We call the 65-template, thirteen-relation setting with annotated boxes the \emph{main setup}.
$t$ and $t+k$ are an early and a late annotated frame, about 3--4\,s apart, and the action is the template.
\emph{RH20T}~\cite{fang2023rh20t} \emph{with RoboInter boxes}~\cite{li2026robointer}: the action is the 6-D end-effector displacement plus the gripper change over $k=3$\,s; whole tasks are held out (112/34 tasks, 98,395/25,104 pairs); the boxes are correct in 34 of 40 hand-checked frames (\cref{sec:rh20t}).
\emph{Tracked objects}: Grounding DINO~\cite{liu2024groundingdino} prompted with ``hand. object.'' on three frames per clip and SAM~2.1~\cite{ravi2025sam2} tracking give up to four tracks, which replace the annotated boxes as the only object input; settings were chosen on training videos.
\emph{Blocks2D}: a 2-D world with gravity, stacking, pushes and lifts and exact relations (\cref{sec:blocks}).
\emph{Training}: frozen ImageNet ResNet-18 unless stated (912,632 trainable parameters for the five-relation model), AdamW at learning rate $2{\times}10^{-4}$, batch 64, and 40 epochs, or a fixed 8,720 steps in the 65-template, 13-relation setup and the experiments built on it; the budget is the same for every arm.
Details are in \cref{sec:impl}.

\section{Experiments}
\label{sec:experiments}

We organise the evidence around four questions.
\textbf{Q1}: does latent prediction ground the relational channels, and does inverse dynamics?
\textbf{Q2}: what makes the channels mean the relations?
\textbf{Q3}: what are grounded channels good for?
\textbf{Q4}: where does grounding stop?

\subsection{Q1: Prediction does not ground the channels; inverse dynamics does}
\label{sec:rq1}

\begin{figure}[t]
\centering
\includegraphics[width=\linewidth]{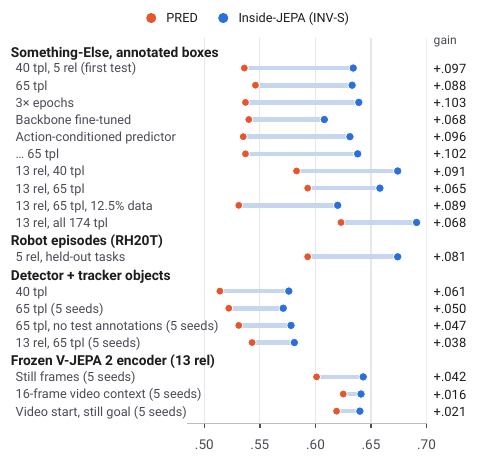}
\caption{\textbf{Inverse dynamics grounds the relational channels across settings.} Matched relational AUROC of \PRED and \InsideJEPA (\INVS), with the gain in each setting; means over 3 seeds, or 5 where marked. tpl: templates; rel: relations.}
\label{fig:grounding}
\end{figure}

\paragraph{A controlled world.}
In Blocks2D, where relations are exact, \PRED reaches .522\,$\pm$\,.018 relational AUROC (mean $\pm$ SD), below an untrained relation head (.641).
Its channels never flip (rate 0.000 against 0.021 in the ground truth), and copying the current channels predicts the next ones as well as the predictor does: it has learned a stable, perfectly predictable code that likely encodes object identity or colour rather than relations.
Inverse dynamics removes this shortcut: .728\,$\pm$\,.009 in the same sweep and .756\,$\pm$\,.008 in a separate matched sweep (\PRED\ .519), and the gain holds on held-out object counts and taller stacks and when vertical lifts rule out a horizontal-motion shortcut (\cref{sec:blocks}).

\paragraph{A first test on real video.}
On Something-Else (40 templates, five relations), \PRED goes wrong in another way: its channels flip at about the true rate (.24 against .21) but do not track the relations, and they score exactly as well as a random relation head (.536, margin +.000).
\INVS reaches .634, a gain of +.097 (about seven seed SDs, pooled over \PRED and \INVS), with a margin gain of +.092 (\twoSE\ .022), and \INVSHUF, trained on shuffled action labels, scores .116 below it (\twoSE\ .016).
A linear probe reads the relations from every arm's pair features (.86--.93), so the information is there even where the channels do not carry it.

\paragraph{Replication.}
With annotated boxes and a ResNet-18 encoder, the gain replicates in all ten settings of \cref{fig:grounding} (top), from +.065 to +.103: on all 65 templates, with three times the epochs, with the backbone fine-tuned (weaker, +.068), with an action-conditioned predictor, on thirteen relations, on 12.5\% of the data and on all 174 Something-Something templates (3$\times$ the pairs).
In every one of these settings the margin gain exceeds \twoSE\ (+.054 to +.095; \cref{tab:annot}).
Trained on steps of about 1\,s instead of frame pairs that span the clip, the gain shrinks to +.037 (\cref{sec:planning}).
It replicates on RH20T, where the action is an end-effector displacement and whole tasks are held out: +.081, with a margin gain of +.084 (\twoSE\ .038); with one camera per setup it is +.046 (five seeds; \cref{sec:rh20t}).
It survives replacing the annotated boxes with detector-and-tracker objects, scored on tracks matched to annotated objects: +.061 on 40 templates and +.052 on 65 (+.050 with two extra seeds).
Scored without any annotation at test time, the same models gain +.047 on 65 templates and +.048 on 40, with the margin gain and the lead over \INVSHUF still beyond \twoSE\ (\cref{sec:tracker}).
The gain is smaller with tracked objects on thirteen relations (+.038) and with a frozen V-JEPA~2 encoder.

\paragraph{All arms.}
\cref{tab:arms} compares the five arms on the main setup.
\INVSHUF falls below \PRED (.539), \INVR barely moves (.608 against .593), and both arms whose inverse head reads the channels with correct action labels ground them: \INVZ (.670) and \INVS (.658).

\begin{table}[t]
\centering
\footnotesize
\setlength{\tabcolsep}{3.2pt}
\begin{tabular}{@{}l cc cccc@{}}
\toprule
 & \multicolumn{2}{c}{Grounding} & \multicolumn{4}{c}{Planning, top-5 of 65} \\
\cmidrule(lr){2-3}\cmidrule(l){4-7}
Arm & \makecell{Matched\\AUROC\up} & Margin\up & \PlChan\up & \PlInv\up & \PlZ\up & \PlS\up \\
\midrule
\PRED    & .593 & +.013 & .067 & \na  & .244 & .168 \\
\INVSHUF & .539 & $-$.026 & .136 & \na  & .237 & .184 \\
\INVR    & .608 & +.020 & .113 & \na  & .311 & .209 \\
\INVZ    & \best{.670} & \best{+.083} & \second{.297} & \best{.482} & \second{.391} & \second{.380} \\
\INVS    & \second{.658} & \second{+.080} & \best{.334} & \second{.374} & \best{.416} & \best{.425} \\
\midrule
\textcolor{ceilinggray}{\SUP} & \textcolor{ceilinggray}{.998} & \na & \textcolor{ceilinggray}{.394} & \na & \na & \textcolor{ceilinggray}{.393} \\
\bottomrule
\end{tabular}
\caption{\textbf{All arms on the main setup} (65 templates, 13 relations, annotated boxes, frozen ResNet-18, 3 seeds; arms as in \cref{fig:arms}). Margin: over a seeded random head. Planning readouts, fitted on relation-labelled pairs: \PlChan, one calibrated channel per relation; \PlInv, the arm's inverse head; \PlZ, a probe on the whole forecast latent $z$; \PlS, a probe on the 16 relational channels $s$ only; prior .174, chance .077. \up: higher is better; bold: best arm trained without relation labels; underline: second best. \SUP is trained on the relation labels and shown only as a ceiling.}
\label{tab:arms}
\end{table}

\subsection{Q2: What grounds the channels}
\label{sec:rq2}

\begin{figure*}[t]
\centering
\includegraphics[width=\linewidth]{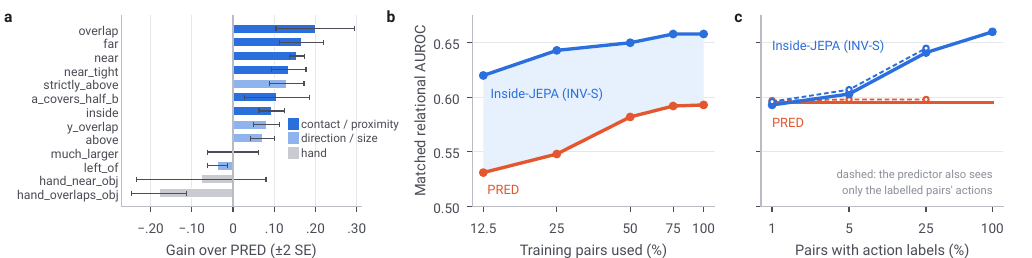}
\caption{\textbf{What drives grounding} (main setup: 65 templates, 13 relations, annotated boxes). \textbf{(a)} Per-relation gain of \INVS over \PRED ($\pm$\twoSE, 3 seeds): contact, proximity and vertical relations ground; \rel{left_of}, \rel{much_larger} and the hand relations do not. \textbf{(b)} Nested fractions of the training data, all labelled (3 seeds): more data lifts \PRED more than \INVS. \textbf{(c)} All data, with action labels on a fraction of the pairs (5 seeds): labels lift \INVS; \PRED has no inverse loss, so its solid line is a single value; dashed: the predictor also sees only the labelled pairs' actions.}
\label{fig:drivers}
\end{figure*}

\paragraph{Decoding the action is not enough.}
\INVR's inverse head reads only the entity latents, so its gradient reaches the channels only through the shared encoder.
On the main setup it decodes the action best of all arms (top-1 .424, against .396 for \INVZ and .316 for \INVS), yet its channels barely ground (.608 \vs\ \PRED's .593; \cref{tab:arms}).
In the setting of the first real-video test and with the action-conditioned predictor, its margin gain over \PRED is +.022 and +.024 (\twoSE\ .016 and .012), beyond noise but about a quarter of \INVS's, and on all 174 templates its channels stay at \PRED's level (.622 \vs\ .623).
What matters is that the inverse head reads the channels, not that it reads only them: \INVZ, which reads the whole latent as recent inverse-dynamics terms in JEPA world models do~\cite{ivashkov2026smwm,zhang2026deltajepa,strohm2026plannable}, grounds about as well as \INVS or better wherever both ran (.654 \vs\ .634 in the first real-video test, .670 \vs\ .658 on the main setup).

\paragraph{The action labels do the grounding.}
No control that keeps the route but replaces the action labels grounds the channels (\cref{tab:controls}).
Shuffled labels never ground ($-$.057).
Two label-free losses on the same channels learn their own tasks (arrow of time .661 against chance .5; InfoNCE top-1 .839 against .016) but do not ground them ($-$.007 and +.010), and both fall short of \INVS by more than \twoSE.
Our reading, which we did not test, is that both tasks can be solved from object appearance and motion without encoding the relations.

\begin{table}[t]
\centering
\footnotesize
\setlength{\tabcolsep}{1.45pt}
\begin{tabular*}{\linewidth}{@{\extracolsep{\fill}}l c c c c c@{}}
\toprule
Loss on $s$ & \makecell{Matched\\AUROC\up} & Gain\up & Effect\up & \PlChan\up & \PlS\up \\
\midrule
None (\PRED) & .595 & \na     & .003 & .066 & .162 \\
Shuffled $a$ (\INVSHUF) & .538 & $-$.057 & .011 & .122 & .186 \\
Arrow of time (\AOT) & .588 & $-$.007 & .046 & .190 & .248 \\
InfoNCE (\NCE) & .605 & +.010   & .006 & .086 & .150 \\
\midrule
$a$ on 1\% (153) & .593 & $-$.002 & .016 & .114 & .222 \\
$a$ on 5\% (701) & .603 & +.008   & .038 & .185 & .280 \\
$a$ on 25\% (3,490) & \second{.641} & \second{+.047} & \second{.089} & \second{.280} & \second{.354} \\
$a$ on all (\INVS) & \best{.660} & \best{+.066} & \best{.103} & \best{.331} & \best{.424} \\
\bottomrule
\end{tabular*}
\caption{\textbf{Grounding needs correct action labels, in the thousands} (65 templates, 13 relations, 5 seeds). Every row trains the same model with a different loss on the relational channels $s$; ``$a$ on $x$\%'': inverse dynamics with action labels on a nested $x$\% of the pairs (number of labelled pairs in brackets), while the predictor trains on all pairs with their true actions; hiding those actions from the predictor too gives the same gains (\cref{tab:fractions}). Gain over \PRED; Effect: action effect on forecasts of changing relations; planning top-5, prior .174. \up: higher is better; bold: best; underline: second best.}
\label{tab:controls}
\end{table}

\paragraph{Grounding needs thousands of correct labels.}
With every pair present but labels on a nested, template-stratified fraction, labels on 1\% or 5\% of the pairs give no measurable grounding gain ($-$.002 and +.008), although at 5\% the action effect and planning already rise (\cref{tab:controls}).
The gain rises between 5\% and 25\%: labels on 25\% of the pairs (3,490, about 54 per template) give +.047 (\twoSE\ .009), against +.066 with all labels, and over all four fractions the gain grows by .011 per doubling (\twoSE\ .002).
Hiding the unlabelled pairs' actions from the predictor too gives the same gains (\cref{sec:labels}).
Labels and data act differently (\cref{fig:drivers}b,c): more training data lifts \PRED from .531 to .593 while \INVS changes less (.620 to .658), whereas more labels lift \INVS and leave \PRED, which has no inverse loss, flat (.597--.598 when it too sees only the labelled pairs' actions).
The thirteen-relation gain shrinks with data only because \PRED learns the two hand relations; the other eleven keep about +.10 from 25\% of the data upward, a split found post hoc and confirmed at two new training-set sizes (\cref{sec:moregrounding}).

\paragraph{Captions or end-effector actions.}
Something-Something templates name the outcome in words, so the inverse loss might be relation supervision in disguise.
A template does not say which channel should carry which relation, and on RH20T, where the action is a 6-D end-effector displacement plus the gripper change, grounding replicates, again carried by contact and proximity (\rel{near} +.22, \rel{overlap} +.10, \rel{above} +.07, \rel{inside} +.05, \rel{left_of} $-$.03).

\paragraph{Grounding is selective.}
The gain is largest for contact and proximity (\cref{fig:drivers}a): \rel{overlap}, \rel{far}, \rel{near} and \rel{near_tight} gain .13--.20, and the vertical relations \rel{strictly_above}, \rel{y_overlap} and \rel{above} gain .07--.13, while \rel{left_of}, \rel{much_larger} and the hand relations gain nothing or lose.
Across datasets and object sources, \rel{overlap} and \rel{near} always gain and \rel{left_of} never does, while the other relations vary (\cref{fig:s_relations}); \rel{inside}, for instance, gains in five of six settings but not on all 174 templates ($-$.046).
A probe reads \rel{left_of} from the same latents (.93 from \PRED's on RH20T).
On Something-Else, how much a relation's change tells about the action predicts its gain (rank correlation +.90 and +.82 on two relation sets), but not on RH20T ($-$.30), so we offer no general account.

\subsection{Q3: What grounded channels are good for}
\label{sec:rq3}

\begin{figure*}[t]
\centering
\includegraphics[width=\linewidth]{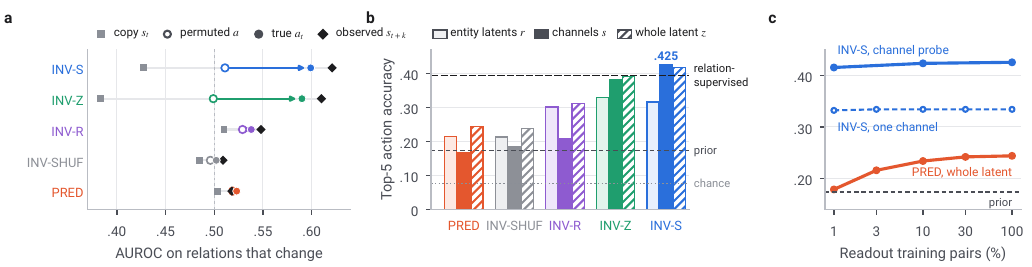}
\caption{\textbf{Using the grounded channels.} \textbf{(a)} Forecasting relational change (40 templates, 5 relations): AUROC on entries whose relation changes, when copying $s_t$, forecasting with a permuted or the true action, or observing $s_{t+k}$; only arms with grounded channels respond strongly to the action. \textbf{(b)} One-step planning from relational goals (top-5 of 65 templates, 13 relations) with probes on the forecast $r$, $s$ or $z$; lines: frequency prior, chance and relation-supervised channels. \textbf{(c)} Relation-labelled pairs needed to fit the readouts.}
\label{fig:use}
\end{figure*}

\paragraph{Forecasting relational change.}
Without the action as input, forecasts of the relations that change stay near chance (.516 for \INVS, .46--.54 across arms).
With it (\cref{fig:use}a), \INVS's forecasts score .599 on the changing entries, close to the channels observed in the future frame (.621), and fall to .511 with a permuted action, an action effect of +.088.
Arms with ungrounded channels barely use the same input (\PRED +.003, \INVR +.009).
The result holds on 65 templates (.579 \vs\ .610 observed) and with tracked objects (.560 \vs\ .569), and when the predictor is rolled forward in steps of about 1\,s the action effect grows from +.021 after one step to +.057 after three (\cref{sec:planning}).
On RH20T, forecasts depend on the action, losing .095 when it is permuted, but reach only .518 (\cref{sec:rh20t}).

\paragraph{Planning from relational goals.}
We test one-step action selection: given the start state and the relations at the end of the clip as the goal, the candidates are ranked by how well each one's forecast matches the goal.
With calibrated one-channel scoring (\PlChan), \INVS ranks the true template in its top five for 33\% of goals, and for 37\% when its inverse head scores the candidates (\PlInv); \PRED, \INVSHUF and \INVR stay below the .174 frequency prior with \PlChan.

\paragraph{The planning signal is in the channels.}
A linear probe on \INVS's 16 relational channels (\PlS) plans at .425 (\cref{fig:use}b), as well as one on its whole latent (.416) and better than one on its entity latents (.316), whereas a probe on \PRED's whole latent reaches .244.
\INVR's channels, trained with the same loss routed around them, plan at .209.
About 140 relation-labelled pairs (1\%) suffice to fit \INVS's readouts (\PlS\ .415 \vs\ .425 with all pairs), whereas \PRED's whole-latent probe falls to .179 (\cref{fig:use}c).
With tracked objects, \PlChan still plans above the prior (.273; .219 with no annotations), although their thirteen-relation gain is smaller (+.038).
The advantage does not depend on the caption's role order either: with annotated boxes (five relations, five seeds), the channel probe beats \PRED's whole latent by +.174 with the objects shuffled and +.175 in caption order, both with the hand last, and by +.116 with the hand first.
Among all 174 templates the one-channel readout falls to the prior, while the inverse head and the channel probe still plan above it (\cref{sec:moregrounding}).
On RH20T, a probe on any arm's latent, \PRED's included, plans best (.213--.229, the best within .008 of a model-free oracle), partly because there is little headroom (\rel{near} holds in 98\% of pairs) and partly because the goals are positional and positions enter every arm's entity latents; we claim relational planning for Something-Else only.

\subsection{Q4: Where grounding stops}
\label{sec:rq4}

\paragraph{Left/right never grounds on real video.}
\rel{left_of} gains in no real-video setting, although a probe reads it from the same latents.
Five explanations that we tested are not supported: captions that name no direction, cameras that disagree on image left/right, no left/right information in the action, the action in the wrong frame and too few channels (\cref{tab:leftof}).
The information is in the latent; it does not become a channel.

\paragraph{The code is distributed.}
One matched channel per relation plans at .334, against .425 for a probe on all 16 channels, and three label-free attempts to give each relation its own channel (a decorrelation penalty, an ICA rotation and a sparsity penalty) made the one-channel readout worse or removed grounding (\cref{sec:planning}).
Matched AUROC is therefore a lower bound on what the channels hold.

\paragraph{A self-supervised video encoder.}
With frozen V-JEPA~2 features~\cite{assran2025vjepa2} on still frames, the gain is +.050 at three seeds and +.042 at five, against +.065 with ResNet-18.
With 16 frames of video context it falls to +.016 (five seeds), as \PRED rises (.601 to .625) while \INVS stays level (.643 to .641); the goal frame's features then reveal the action through motion (an action probe on them rises from .479 to .566), and a still goal frame restores the inverse head's accuracy but not the gain (+.021).
With still frames, by contrast, \PRED barely changes from ResNet-18 (.599 \vs\ .593) and \INVS is lower (.648 \vs\ .658), so the smaller gain has no single cause.
Channel-probe planning holds throughout (.390--.399 \vs\ .252--.308 for \PRED).
V-JEPA~2 saw 87\% of our validation clips in pretraining; on the 1,918 it did not see, the margin gain is unchanged (+.042 \vs\ +.045 on size-matched seen clips; \cref{sec:vjepa}).

\paragraph{Distance to a supervised ceiling.}
Channels trained on the relation labels reach .998, as the relations are near-exact functions of the box coordinates.
Inverse dynamics recovers .16 of that ceiling's gain over \PRED with one-channel matching and .37 with a channel probe (V-JEPA~2: .12 and .25): about a third for contact and proximity with matching and about half with a probe, and little or none for \rel{left_of}, \rel{much_larger} and the hand relations.
Yet the inverse-trained channels plan as well as the supervised ones through a probe (.424 \vs\ .393), although the supervised ones show about twice the action effect when forecasting relational change (.223 \vs\ .103; \cref{sec:ceiling}).

\section{Discussion and limitations}
\label{sec:discussion}

\paragraph{What is shown, and what is not.}
With architecture, data and budget fixed, inverse dynamics on the relational channels aligns them with relations that latent prediction leaves them free to ignore, on caption-labelled video, on robot episodes with end-effector actions and, more weakly, with tracked objects.
This needs an inverse head that reads the channels and correct action labels on thousands of pairs.
Grounding here means that one channel per relation scores above \PRED and above a random relation head against relations computed from boxes; it does not show that the channels represent physical relations beyond what the boxes define.
Nor do the experiments show why particular relations ground: our design rationale (\cref{sec:inverse}) fits Something-Else, where the relations whose change reveals the action gain most, but not RH20T.
The two label-free objectives we tried learn their tasks without grounding the channels; that does not rule out every label-free objective.

\paragraph{Why it matters for world models.}
A world model trained for control receives actions anyway.
Our results suggest that making relational channels explain those actions is a practical way to give them meaning, and that doing so turns them into a compact readout for stating goals (\cref{fig:use}b).
Recent JEPA world models already add inverse-dynamics terms on the whole latent~\cite{ivashkov2026smwm,boylan2026acmtm,zhang2026deltajepa,strohm2026plannable}; our \INVZ arm suggests that such terms also ground relational channels placed inside the state, as long as the inverse head reads them.

\paragraph{Limitations.}
\emph{Scope}: a controlled study against one baseline, \PRED, with the architecture held fixed; we ran no other relational or inverse-dynamics method, and \INVZ stands in for whole-latent inverse dynamics.
\emph{Data and horizons}: one human-video dataset with annotated boxes and one robot dataset, whose boxes are correct in 34 of 40 hand-checked frames; the Something-Something action is a caption; horizons are within clips of about 4\,s.
\emph{Objects}: with tracked objects the effect is half to two-thirds as large (+.061 and +.052 on 40 and 65 templates, against +.096 and +.102 with annotated boxes); missed objects explain most of the gap, and the caption's role order, encoded in Something-Else's entity order but unknown to a tracker, part of the rest (\cref{sec:tracker}); on RH20T, automatically found boxes were correct in at most 18 of 40 hand-checked frames, so the robot result rests on the given boxes.
\emph{Effect size}: gains of .02--.10 and .12--.37 of a supervised ceiling are modest, although the test is strict (one channel per relation, against an identical architecture), and the channels form a distributed code, not clean predicates.
\emph{Encoders and training}: frozen backbones except one fine-tuned run, where the effect is weaker; with V-JEPA~2 the gain is smaller (+.042); one budget and loss weight; three to five seeds.
\emph{Planning}: one-step action selection within a clip, with readouts fitted on relation-labelled pairs, on Something-Else only.

\section{Conclusion}
\label{sec:conclusion}
Relational channels inside a JEPA world model do not become meaningful by being there: latent prediction leaves them free to mean nothing.
Making them explain the action aligns them with the relations, on caption-labelled video, on robot episodes and with automatically found objects, provided the inverse head reads the channels and enough actions are labelled.
The grounded channels forecast how actions change relations and carry the signal for selecting actions towards relational goals: selectively, as a distributed code, and at a level that a stronger encoder lets the plain predictor approach.
We hope that the controlled comparisons, and the negative results reported beside the positive ones, make \InsideJEPA a sound starting point for world models whose states can be read and given goals in relational terms.

{
    \small
    \bibliographystyle{ieeenat_fullname}
    \bibliography{references}
}

\appendix
\maketitlesupplementary
\noindent The appendices give implementation and evaluation details (\cref{sec:impl,sec:evaldetails}); further grounding results (\cref{sec:moregrounding}); the controlled 2-D world (\cref{sec:blocks}); objects from a detector and tracker (\cref{sec:tracker}); the robot episodes (\cref{sec:rh20t}); label-free objectives and label fractions (\cref{sec:labels}); planning and the distributed code (\cref{sec:planning}); the V-JEPA~2 encoder (\cref{sec:vjepa}); the supervised ceiling (\cref{sec:ceiling}); and the decision rules \Crule{1}--\Crule{6}.

\section{Implementation details}
\label{sec:impl}

\paragraph{Model.}
Entities are the objects and hands annotated (or tracked) in both frames of a pair, at least two and at most four; frames with fewer are padded and frames with more are truncated to four.
Each entity box is cropped from the frame, resized to $96\times96$ pixels without keeping its aspect ratio, and encoded by a frozen ImageNet ResNet-18 with ImageNet input normalisation into one average-pooled 512-d vector. $f$ adds a learned linear projection of this vector, an MLP of the box's Fourier features ($\gamma$ in \cref{eq:entity}: sines and cosines of $2^k\pi$ times each normalised coordinate, $k=0,\dots,3$, plus the coordinates) and a learned embedding of the entity's type (object or hand), and applies LayerNorm; so the box coordinates enter every entity embedding, and an untrained relation head already sees geometry.
Entity embeddings have 128 dimensions and entity latents 32; the relation head $g$, an MLP with one hidden layer of 256 units on $(e_i, e_j, e_i-e_j, e_i\odot e_j)$, produces $K=8$ (five relations) or $K=16$ (thirteen relations) sigmoid channels per ordered pair, and pair entries for padded entities and for $i=j$ are held constant.
With $K=8$ the flattened state has 256 dimensions, and the action-conditioned predictor is \texttt{Linear(256+64$\to$512)}, GELU, \texttt{Linear(512$\to$256)} applied to the state and a 64-d learned embedding of the template; the first four settings of \cref{tab:annot} and the test of forecasting without the action use the same predictor without the action input.
On RH20T the control vector enters through a linear layer.
The target encoder is an exponential moving average of the online encoder (momentum 0.996).
The inverse head reads $s_{t+k}$ from the online encoder.
With the frozen ResNet-18 the five-relation model has 912,632 trainable parameters; fine-tuning the backbone raises this to 12,241,248.

\paragraph{Training.}
One AdamW optimiser (learning rate $2{\times}10^{-4}$, weight decay $10^{-4}$), batch 64, $\lambda_{\text{inv}}=1$, identical for every arm and seed.
Budgets were fixed before any result from the training loss of one pilot run (the smallest multiple of 10 epochs within 0.1 of the 60-epoch value): 40 epochs in the experiments with five relations or 40 templates (120 in the run with three times the epochs) and in the robot and tracker experiments, and a fixed 8,720 steps (about 40 epochs of the full training set) in the 65-template, 13-relation setup and the experiments built on it, so that the budget does not change across data and label fractions.
Seeds 0--2, plus 3--4 whenever a locked seed rule (\Crule{6} in \cref{sec:registry}, or its analogue) fired or, where marked, post hoc; one experiment used five seeds by design.
All intervals are over training seeds, and each states whether it is a standard deviation (SD), a standard error (SE) or two standard errors (\twoSE) of a mean or of a seed-paired difference; with three to five seeds, \twoSE is a lenient interval, and it reflects seed variance only.

\paragraph{Frames and horizons.}
Something-Something V2 frames are extracted at the native 12\,fps and are pixel-identical to the official Something-Else sample frames.
$t$ and $t+k$ are an early and a late annotated frame of the clip, about 3--4\,s apart.
On RH20T, $k$ is 30 frames (3\,s at 10\,Hz).



\section{Evaluation details}
\label{sec:evaldetails}

\begin{figure*}[t]
\centering
\includegraphics[width=\textwidth]{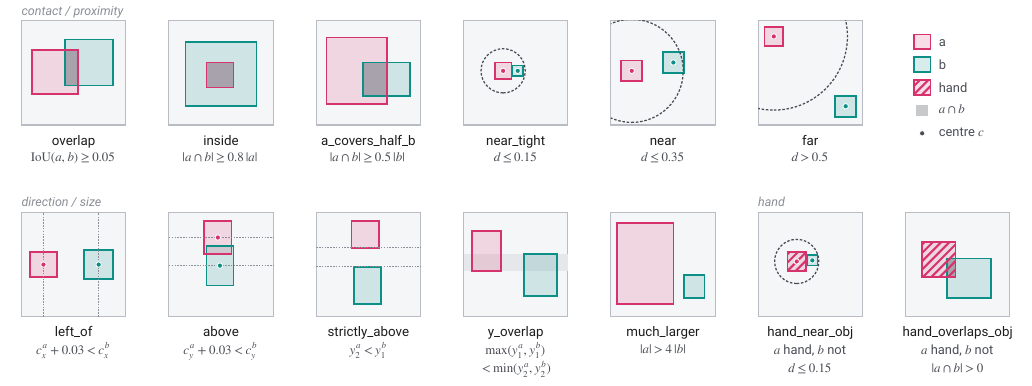}
\caption{\textbf{The thirteen relations}, each with one example frame of a directed pair $a\to b$ (rose, teal; hatched: a hand) that satisfies it. Boxes are normalised to $[0,1]$ with $y$ pointing down. Under each name is its rule as written in the code. $c$: box centre; $|\cdot|$: area; $d=\lVert c^a-c^b\rVert/\sqrt{2}$. A dashed ring marks the $d$ threshold around $c^a$; dark grey marks $a\cap b$; the light-grey band in \textsf{y\_overlap} is the shared vertical extent; dotted lines mark the compared centres or edges.}
\label{fig:s_reldefs}
\end{figure*}

\paragraph{Relations.}
All relations are computed from the entity boxes of a frame and are used only for evaluation (and as targets for the \SUP ceiling).
Each box is clamped to the frame and divided by the frame's width and height, so $(x_1,y_1,x_2,y_2)\in[0,1]^4$ with $y$ pointing down.
Every ordered pair $(a,b)$ of distinct entities gets its own relation vector, from the box edges, centres $c^a,c^b$, areas $|a|,|b|$, intersection $|a\cap b|$ and centre distance $d=\lVert c^a-c^b\rVert/\sqrt{2}\in[0,1]$, and, for the hand relations, which entity is the hand.
\Cref{fig:s_reldefs} gives each rule as written in the code: pairs with $0.35<d\le0.5$ are neither \rel{near} nor \rel{far}, and both hand relations are 0 unless $a$ is a hand and $b$ is not.
The code computes in 32-bit floats with a $10^{-8}$ guard in each ratio, so a pair lying exactly on a threshold can fall on either side of it.
The five-relation set is \rel{left_of}, \rel{above}, \rel{overlap}, \rel{inside}, \rel{near}; the thirteen-relation set adds \rel{far}, \rel{near_tight}, \rel{y_overlap}, \rel{a_covers_half_b}, \rel{much_larger}, \rel{strictly_above}, \rel{hand_near_obj} and \rel{hand_overlaps_obj}.
For grouping we call \rel{overlap}, \rel{near}, \rel{near_tight}, \rel{far}, \rel{inside} and \rel{a_covers_half_b} \emph{contact and proximity}; \rel{left_of}, \rel{above}, \rel{strictly_above}, \rel{y_overlap} and \rel{much_larger} \emph{direction and size}; and the two hand relations \emph{hand}.

\paragraph{Matched relational AUROC and margin.}
\cref{eq:mauroc} is computed on the relational channels of the validation pairs over all valid ordered pairs, with one channel per relation chosen by Hungarian matching on those same pairs.
Fitting the matching on training pairs instead leaves \PlChan unchanged (at most $-$.015, with labels on 1\% of the pairs).
The random-head reference replaces the trained relation head $g$ with a freshly initialised one (seed 0), applied to the checkpoint's own entity embeddings and evaluated identically; with box coordinates in the input it scores .52--.60 per relation.
The probe is a linear classifier per relation on each arm's pair features; \Crule{4} requires probe AUROC $\ge .75$ in every arm.

\paragraph{Forecasting metrics.}
\emph{m1} is the copy error divided by the predictor error on the channel entries of valid pairs; \emph{m2} is the matched AUROC of the forecast $\hat s_{t+k}$ against the relations at $t+k$, using each run's grounding matching; \emph{m3} is m2 restricted to entries whose relation changes between $t$ and $t+k$ (the metric of \cref{fig:use}a).
References: copying $s_t$, the channels of the observed future frame, and a forecast with the action permuted across the validation set by a fixed permutation.
The action effect \emph{m4} is m3 with the true action minus m3 with the permuted one.

\paragraph{Planners.}
Candidates are all templates (40, 65 or 174); the rank of the true action is one plus the number of candidates with a strictly better score.
References are chance (5/65 = .077), a template-frequency prior (.174 on 65 templates) and a model-free oracle that compares each template's mean ground-truth relation change in training with the pair's actual change.
\PlDist scores a candidate by the squared distance between forecast matched channels and goal relations.
\PlChan scores the likelihood of the goal relations under per-relation logistic calibrations of the forecast matched channels.
\PlInv (amortised inverse) maps the goal relations into channel space through per-channel class means of the matched channels, fitted on training pairs, and gives the start channels and this goal to the arm's inverse head; candidates are ranked by its output.
\PlZ, \PlS and \PlR score the goal under per-relation linear logistic probes on the forecast pair features $(r_i, r_j, s_{ij})$, on $s_{ij}$ only, or on $(r_i, r_j)$ only.
Readouts are fitted on labelled training pairs; about 140 (1\%) suffice for \PlChan and \PlInv, and \PlZ gains +.030 from 1\% to 100\% for \INVS.
On RH20T, whose inverse head regresses a control, \PlInv ranks the 40 candidate controls by their distance to its output.

\paragraph{Why calibrated planners.}
With five relations, grounded arms planned better than ungrounded ones on 40 templates (\INVS\ .234 \vs\ \PRED\ .130; prior .194, oracle .298) but did not beat the prior on 65 templates, and with thirteen relations even the 40-template margin over the prior fell just inside \twoSE\ (.228 \vs\ .194).
An evaluation-only analysis then showed that raw distance scoring was the cause (calibrated \PlChan\ .363 and \PlInv\ .456 on the 40-template, thirteen-relation checkpoints), and the calibrated planners were locked before the 65-template replication.

\section{Additional grounding results}
\label{sec:moregrounding}

\paragraph{Settings with annotated boxes.}
\cref{tab:annot} gives the values behind the top block of \cref{fig:grounding} with the seeded \Crule{2} margin gains.
Gains are computed from unrounded per-seed AUROCs, so they can differ by .001 from the rounded columns.

\begin{table}[!htb]
\centering
\scriptsize
\setlength{\tabcolsep}{1.7pt}
\begin{tabular}{@{}l cc ccc c@{}}
\toprule
Setting & Tpl & Rel & \PRED\up & \INVS\up & Gain\up & \makecell{Margin gain\\(\twoSE)} \\
\midrule
First test & 40 & 5 & .536 & .634 & +.097 & +.092 (.022) \\
All templates & 65 & 5 & .546 & .633 & +.088 & +.084 (.021) \\
3$\times$ epochs & 40 & 5 & .537 & .639 & +.103 & +.089 (.026) \\
Fine-tuned backbone & 40 & 5 & .540 & .608 & +.068 & +.080 (.034) \\
Action-cond.\ $P$ & 40 & 5 & .535 & .631 & +.096 & +.095 (.013) \\
\quad all templates & 65 & 5 & .537 & .638 & +.102 & +.091 (.040) \\
Thirteen relations & 40 & 13 & .583 & .674 & +.091 & +.086 (.018) \\
\quad all templates & 65 & 13 & .593 & .658 & +.065 & +.067 (.021) \\
\quad 12.5\% of the data & 65 & 13 & .531 & .620 & +.089 & +.084 (.019) \\
\quad all SSv2 templates & 174 & 13 & .623 & .691 & +.068 & +.054 (.015) \\
\bottomrule
\end{tabular}
\caption{\textbf{Grounding with annotated boxes}, frozen ResNet-18 (fine-tuned in one row), 3 seeds. Every row passes \Crule{1}--\Crule{6} except 3$\times$ epochs (120), which trained only \PRED and \INVS and passes its own rule that the gap persists. Margin gain: \INVS's margin over the seeded random head minus \PRED's. The first four rows use $P(z_t)$; subsequent rows use $P(z_t, u(a_t)$.}
\label{tab:annot}
\end{table}

\paragraph{The first test.}
\cref{tab:v1} gives every arm of the first test on real video.
\Crule{1}: +.097 (bar .05).
\Crule{2}: margin gain +.092 (\twoSE\ .022) with the seeded random head (+.116, \twoSE\ .028, with the original unseeded random head).
\Crule{3}: \INVS $-$ \INVSHUF +.116 (\twoSE\ .016).
\Crule{4}: probes .86--.93.
\Crule{6}: the gain is about seven pooled seed SDs, so no extra seeds.
\PRED's channels flip about as often as the ground truth but do not track it, while \INVSHUF's flip most and track least.
\cref{tab:v6} gives every arm with the action-conditioned predictor on all 65 templates, including forecasting.

\begin{table}[!htb]
\centering
\scriptsize
\setlength{\tabcolsep}{2.4pt}
\begin{tabular}{@{}l cccc c c@{}}
\toprule
Arm & \makecell{Matched\\AUROC\up} & \makecell{Random\\head} & Margin\up & Probe\up & \makecell{Flip rate\\(GT .212)} & \makecell{Own inverse\\top-1\up} \\
\midrule
\PRED & .536 & .536 & +.000 & .882 & .242 & \na \\
\INVSHUF & .517 & .535 & $-$.017 & .856 & .378 & .025 \\
\INVZ & \best{.654} & .535 & \best{+.119} & \second{.914} & .324 & \best{.409} \\
\INVS & \second{.634} & .542 & \second{+.092} & \best{.930} & .302 & \second{.308} \\
\bottomrule
\end{tabular}
\caption{\textbf{First test on real video} (40 templates, 5 relations, compositional validation, 3 seeds). Random head and margin are seeded; inverse top-1 chance is .025. \up: higher is better; bold: best; underline: second best.}
\label{tab:v1}
\end{table}

\begin{table}[!htb]
\centering
\scriptsize
\setlength{\tabcolsep}{2.2pt}
\begin{tabular}{@{}l c c c cccc c@{}}
\toprule
Arm & \makecell{Matched\\AUROC\up} & Margin\up & Probe\up & \makecell{m3\\forecast\up} & Copy & \makecell{Obs.} & Perm. & \makecell{m4\up} \\
\midrule
\PRED & .537 & +.006 & .896 & .513 & .506 & .514 & .512 & +.001 \\
\INVSHUF & .523 & +.002 & .856 & .500 & \na & \na & \na & +.002 \\
\INVR & .551 & +.014 & .919 & .505 & \na & \na & \na & +.008 \\
\INVZ & \second{.635} & \best{+.097} & \second{.930} & \best{.593} & .378 & .620 & .485 & \best{+.108} \\
\INVS & \best{.638} & \second{+.096} & \best{.938} & \second{.579} & .400 & .610 & .506 & \second{+.074} \\
\bottomrule
\end{tabular}
\caption{\textbf{Action-conditioned predictor on all 65 templates} (5 relations, 3 seeds). m3: AUROC on changing entries with the true action; Copy: copying $s_t$; Obs.: observed future channels; Perm.: permuted action; m4: action effect. Seeded margins; \Crule{2} margin gain +.091 (\twoSE\ .040). \up: higher is better; bold: best; underline: second best.}
\label{tab:v6}
\end{table}

\paragraph{The inverse head on the entity latents only.}
Margins over the seeded random head, 3 seeds, in five settings (first-test setting, 65 templates, fine-tuned, action-conditioned, action-conditioned on 65 templates): \INVR +.023, +.036, +.011, +.021, +.014; \INVS +.092, +.094, +.074, +.092, +.096; \INVZ +.119, +.103, +.102, +.093, +.097. \INVR was added after the first test, in a separate run of the same setting.
\INVR's margin gain over \PRED is +.022 (\twoSE\ .016) in the first test's setting and +.024 (\twoSE\ .012) with the action-conditioned predictor: small, but beyond \twoSE once the random head is seeded.
On the main setup, \INVR's own inverse head is the most accurate of all arms (top-1 .424 from the entity latents, against .316 for \INVS from the channels and .396 for \INVZ), yet its channels barely ground (.608 \vs\ \PRED's .593): recovering the action is not enough; for strong grounding the inverse head has to read the channels.

\begin{figure}[!htb]
\centering
\includegraphics[width=\linewidth]{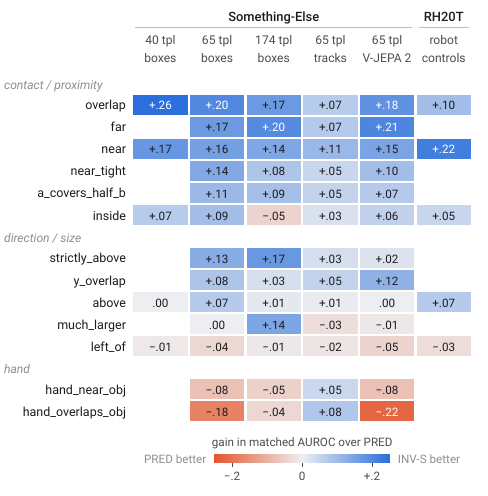}
\caption{\textbf{Per-relation gain} of \INVS over \PRED in matched AUROC: annotated boxes on 40, 65 and all 174 templates, tracked objects and a frozen V-JEPA~2 encoder on 65 templates, and RH20T robot episodes. 3 seeds (V-JEPA~2: 5); a blank cell is a relation the setting did not evaluate.}
\label{fig:s_relations}
\end{figure}

\begin{figure}[!htb]
\centering
\includegraphics[width=\linewidth]{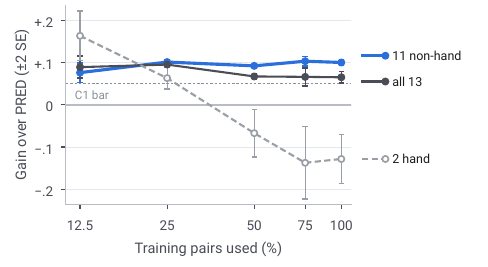}
\caption{\textbf{Grounding gain by training-set size} for the eleven non-hand relations, all thirteen and the two hand relations (65 templates, all pairs labelled, 3 seeds, $\pm$\twoSE). Dashed: the \Crule{1} bar.}
\label{fig:s_datascale}
\end{figure}

\paragraph{Per relation.}
\cref{tab:perrel} gives the per-relation values of the first test: the whole gain comes from \rel{overlap}, \rel{near} and \rel{inside}, while \rel{left_of} and \rel{above} do not beat \PRED or the random head in any arm, although the probe decodes them at about .93.
On thirteen relations and 40 templates the largest gains are \rel{overlap} +.215, \rel{near} +.184, \rel{far} +.184 and \rel{near_tight} +.119, while \rel{left_of}, \rel{much_larger} and \rel{hand_overlaps_obj} gain nothing.
On 65 templates (\cref{fig:drivers}a), the eleven non-hand relations gain +.100 $\pm$ .008 together and the two hand relations $-$.128 $\pm$ .088 (\twoSE over seeds, unpaired; \cref{fig:s_datascale} uses seed-paired intervals).
On RH20T: \rel{near} +.22, \rel{overlap} +.10, \rel{above} +.07, \rel{inside} +.05, \rel{left_of} $-$.03.
\cref{fig:s_relations} sets six settings side by side: \rel{overlap} and \rel{near} gain in every setting and \rel{far} wherever it is scored, \rel{left_of} in none, and \rel{inside} in five of the six (not on 174 templates).

\begin{table}[!htb]
\centering
\scriptsize
\setlength{\tabcolsep}{3.5pt}
\begin{tabular}{@{}l c cccc@{}}
\toprule
Relation & Base rate & \PRED & \INVS & \INVSHUF & \INVZ \\
\midrule
\rel{left_of} & .445 & \best{.55} & .54 & .51 & \best{.55} \\
\rel{above}   & .443 & {.53} & {.53} & .52 & \best{.55} \\
\rel{overlap} & .432 & .51 & \best{.77} & .53 & \best{.77} \\
\rel{inside}  & .093 & .55 & {.62} & .51 & \best{.72} \\
\rel{near}    & .745 & .54 & \best{.71} & .52 & {.67} \\
\bottomrule
\end{tabular}
\caption{\textbf{Per-relation matched AUROC} in the first test (validation base rates). The random head scores .52--.60 and the probe .80--.95 on every relation. Bold: best arm per relation.}
\label{tab:perrel}
\end{table}

\paragraph{Data scale.}
No larger annotated set exists for these templates, so we trained on nested subsets (\cref{tab:datascale}).
Grounding passes at every size.
Over all five sizes the thirteen-relation gain shrinks by .010 per doubling (\twoSE\ .005; .015 over the first three, 25--100\%), only because \PRED learns the two hand relations: the hand slope is $-$.105 (\twoSE\ .005) per doubling and the non-hand slope +.011 (\twoSE\ .012) from 12.5\% to 75\%.
\cref{fig:s_datascale} shows the split.

\begin{table}[!htb]
\centering
\scriptsize
\setlength{\tabcolsep}{3pt}
\begin{tabular}{@{}l r ccc cc@{}}
\toprule
Data & Pairs & All 13\up & 11 non-hand\up & 2 hand\up & \PRED & \INVS \\
\midrule
12.5\% & 1,746 & +.089 & +.076 & +.163 & .531 & .620 \\
25\% & 3,490 & +.095 & +.101 & +.063 & .548 & .643 \\
50\% & 6,980 & +.067 & +.092 & $-$.067 & .582 & .650 \\
75\% & 10,464 & +.066 & +.103 & $-$.137 & .592 & .658 \\
100\% & 13,954 & +.065 & +.100 & $-$.128 & .593 & .658 \\
\bottomrule
\end{tabular}
\caption{\textbf{Grounding gain by training-set size} (65 templates, 13 relations, all pairs labelled, 3 seeds); the 12.5\% and 75\% rows were added later, under rules locked to test the hand/non-hand split.}
\label{tab:datascale}
\end{table}

\paragraph{All 174 templates.}
This setting trains on every template of the compositional split (174 templates; 42,164 training and 44,342 validation pairs, 3.0$\times$ the 65-template setup) for the same 8,720 steps.
Every rule passes: \Crule{1} +.068; \Crule{2} margin gain +.054 (\twoSE\ .015; margins \INVS +.092, \PRED +.038); \Crule{3} \INVS $-$ \INVSHUF +.128; \Crule{4} probes .90--.97; \Crule{5} passes; \Crule{6} does not fire, so three seeds are final.
\INVZ again grounds most (.712, margin +.111) and \INVSHUF least (.563, $-$.006), while \INVR falls to \PRED's level (.622 \vs\ .623; margin +.006 \vs\ +.038); \INVS reaches .691.
A secondary rule, hashed with the others, required the eleven non-hand relations to gain at least .05: they gain +.088 (+.100 on 65 templates), and the two hand relations $-$.045.
Per relation (\cref{fig:s_relations}), \rel{much_larger} and \rel{strictly_above} gain more than on 65 templates (+.143 and +.169), \rel{inside} turns negative ($-$.046), and \rel{left_of} again does not ground ($-$.013).
Planning among 174 templates (top-5 chance .029, frequency prior .077) is not comparable with \cref{tab:planning}: \INVS's amortised inverse \PlInv reaches .341 and its channel probe \PlS\ .179, against .060 for \PRED's, while the one-channel readout \PlChan falls to the prior (.081).

\section{Blocks2D}
\label{sec:blocks}

\paragraph{Environment.}
A 2-D world with gravity, stacking, horizontal pushes and carried blocks, rendered to RGB.
Binary relations \rel{on}, \rel{touching}, \rel{left_of} and \rel{above} (and unary \rel{moving}, \rel{on_ground}, \rel{at_wall}) are exact.
Splits: IID, held-out object counts (6--7 objects) and forced taller stacks.
Base rates: \rel{on} .20, \rel{touching} .43, \rel{moving} .39.
Objects enter through an oracle RoI encoder with raw box coordinates in the entity latent, as on real video.

\paragraph{Results.}
A prediction-only relation head scores .522 $\pm$ .018 (mean $\pm$ SD over 3 seeds) IID binary AUROC against an untrained head at .641, and its channels never flip (rate 0.0000 against a ground-truth 0.0206); copying the current channels matches the predictor.
In the first full run with inverse dynamics, binary AUROC is .758 IID (random head .652), .738 on held-out counts and .775 on tall stacks; flips rise to .042 and the predictor beats copying by about 2.2$\times$.
Per relation (IID): \rel{left_of} .930, \rel{above} .836, \rel{on} .749, \rel{touching} $\approx$ .518.
A separate matched sweep of 20k steps gives .756 $\pm$ .008 against .519 $\pm$ .004 for prediction only.
Unlike real video, \rel{left_of} and \rel{above} ground in Blocks2D, where pushes are horizontal, and they still do when vertical lifts are added.

\paragraph{Vertical actions.}
Because pushes are horizontal, inverse dynamics could exploit $x$-motion and make \rel{left_of} artificially easy.
Adding vertical lifts (17 to 25 actions) leaves grounding unchanged (IID .757 $\pm$ .010 horizontal-only, .755 $\pm$ .016 with lifts; tall stacks .763 and .759); on tall stacks \rel{above} rises (.852 to .889) and \rel{left_of} falls within its seed variation (.920 to .885).

\begin{table}[!htb]
\centering
\scriptsize
\begin{tabular}{@{}l c@{}}
\toprule
Arm & IID binary AUROC\up \\
\midrule
Prediction only & .522 $\pm$ .018 \\
+ inverse dynamics & .728 $\pm$ .009 \\
+ inverse dynamics + unary inverse head & {.743} $\pm$ .011 \\
+ inverse dynamics, no residual bottleneck$^{}$ & {.738} $\pm$ .005 \\
+ inverse dynamics + object masking & .651 $\pm$ .023 \\
\bottomrule
\end{tabular}
\caption{\textbf{Blocks2D ablations} (mean $\pm$ SD over 3 seeds; untrained head .641;).}
\label{tab:blocks}
\end{table}

\section{Detector-and-tracker objects}
\label{sec:tracker}

\paragraph{Pipeline.}
Grounding DINO (tiny) with the prompt ``hand. object.'', box threshold .20, detects on the first, middle and last frame of each clip, with non-maximum suppression and a new-track rule for later detections; SAM~2.1 (hiera-small) tracks each proposal through the clip; up to four tracks are kept, hand-labelled tracks first.
Track boxes replace the annotated boxes as the model's only object input.
Settings were selected on a 500-video development sample from the training set by coverage, the share of annotated entities matched by a track at IoU $\geq$ .1 in both frames of a pair (.853 for the chosen setting, the highest).

\paragraph{Object sources.}
Slots failed first (\cref{tab:objsources}): DINOSAUR slots on DINOv2 features~\cite{seitzer2023dinosaur,oquab2024dinov2} decode the relations (probe .88) but do not keep an object in the same slot over a clip (identity .71), and grounding is +.018; three further slot or mask variants failed pre-registered identity or coverage gates.

\begin{table}[!htb]
\centering
\scriptsize
\setlength{\tabcolsep}{2.4pt}
\begin{tabular}{@{}l c c l@{}}
\toprule
Objects & Identity\up & Found\up & Outcome \\
\midrule
DINOSAUR slots on DINOv2 & .71 & 59\% & +.018, fails \\
Slots re-ordered post hoc & .72 & \na & identity gate fails \\
SAVi-style tracking slots & .53 & \na & identity gate fails \\
SAM~2 automatic masks & \best{.96} & 34\% & coverage gate fails \\
GDINO + SAM~2, 1 frame & {.94} & {64\%} & {+.041}, fails L1 \\
GDINO + SAM~2, 3 frames & .87 & \best{79\%} & \best{+.061, pass} \\
Larger GDINO + SAM~2 & .87 & 50\% & gate fails \\
\bottomrule
\end{tabular}
\caption{\textbf{Object sources tried.} GDINO: Grounding DINO. Identity: the same object stays in the same slot or track over the clip; Found: annotated objects matched at IoU $\ge .5$. \up: higher is better; bold: best.}
\label{tab:objsources}
\end{table}

\paragraph{Scoring.}
Each track is matched per frame to an annotated entity by Hungarian matching on box IoU, and only matched pairs are scored; this uses annotations at test time and is an upper bound.
The annotation-free score instead computes the relations from the tracks' own boxes and scores all track pairs; pairs involving distractor tracks ground about as well as real-object pairs (+.045 \vs\ +.049).
The rules L1--L4 of these experiments correspond to \Crule{1}--\Crule{4}.
\cref{tab:tracker} summarises every test with tracked objects.

\begin{table}[!htb]
\centering
\scriptsize
\setlength{\tabcolsep}{2.5pt}
\begin{tabularx}{\linewidth}{@{}l X@{}}
\toprule
Test & Tracked objects (annotated boxes) \\
\midrule
Grounding, 40 tpl, 5 rel & +.061, pass (+.096) \\
\quad no annotations at test & +.048, L1 fails (L2--L4 hold) \\
Grounding, 65 tpl, 5 rel & +.052, pass; post hoc 5 seeds +.050 (+.102) \\
\quad no annotations at test, 5 seeds & +.047, L1 fails (L2--L4 hold) \\
Grounding, 65 tpl, 13 rel & +.043, below the bar; post hoc 5 seeds +.038 (+.065) \\
Forecast, true / permuted action, 5 seeds & .560 / .495 (.579 / .506) \\
\quad observed future & .569 (.610) \\
Planning, 13 rel, \PlChan, 5 seeds & .273 (.334); no annotations .219 \\
\PlS \vs\ \INVR's \PlZ & .334 \vs\ .288 (.425 \vs\ .311) \\
\bottomrule
\end{tabularx}
\caption{\textbf{Tracked objects}: no box annotations in training; 3 seeds unless marked (marks refer to tracked objects). Values for annotated boxes in brackets. Planning: top-5 among 65 templates, prior .174. The forecasting action effect is 86\% of the annotated-box one.}
\label{tab:tracker}
\end{table}

\begin{figure}[!htb]
\centering
\includegraphics[width=\linewidth]{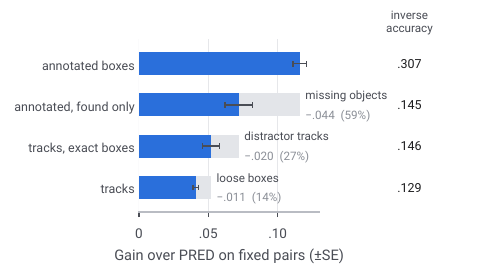}
\caption{\textbf{Where the gap to annotated boxes comes from}, on one fixed set of validation pairs (gain of \INVS over \PRED, 5 seeds, $\pm$SE). From top: all annotated boxes; annotated boxes for the objects the tracker (detection on one frame) found; its tracks with boxes snapped to the annotations; its tracks as run. Each pale segment is what one step costs. Right: the inverse head's action accuracy. Seeds 3--4 were added after the locked three-seed reading.}
\label{fig:s_tracker}
\end{figure}

\paragraph{The gap to annotated boxes.}
For the one-frame tracker, on a fixed set of validation pairs, the locked three-seed reading found the gap spread over several causes; with five seeds (post hoc, \cref{fig:s_tracker}), missed objects account for +.044 $\pm$ .013 of a +.075 gap (59\%), distractor tracks for +.020 $\pm$ .013 (27\%), and loose boxes for +.011 $\pm$ .004 (14\%).
Of the 5,497 annotated entities (objects and hands) still missed on 65-template validation, 3,080 were detected but not among the four tracks kept, 1,209 were never detected and 1,208 were lost in tracking; keeping a fifth track recovers objects but adds as many distractors (42\% to 52\% of kept tracks) and does not change the gain (+.083 \vs\ +.080); choosing tracks by proximity to the hand did not reduce distractors on validation.
Larger models do not recover them: with Grounding DINO base and SAM~2.1 large, the same prompt and the same threshold rule, 50.4\% of annotated validation entities get a track at IoU $\ge .5$ at frame $t$ (78.1\% with the small models on the same entities), short of the locked 81.1\% bar, so nothing was trained.
Swapping one model at a time puts the loss on the larger detector, which scores the generic ``object'' much lower (median top score .20 \vs\ .47), while the larger tracker changes nothing.
Part of the annotated-box advantage is a property of the \PRED baseline: in the five-relation runs with the three-frame tracker, \PRED's channels score below its random head (margins $-$.020 to $-$.036) while \INVS's margin stays at +.035 to +.039 (margins as first evaluated, before the seeded re-score, which moved these runs' margins by about .01 per checkpoint and changed no verdict; \cref{sec:registry}).
The shuffled-label control, which behaves like \PRED, is the cleaner comparison: the true action lifts the channels by .045--.059 above it.

\paragraph{Input format and role order.}
Crossing training and test objects on the same validation pairs (\cref{tab:format}) shows that, in the tracker's format, neither training nor testing on annotated boxes moves the gain beyond \twoSE.
The hand flag adds nothing; listing the hand last helps with annotated boxes but not with tracked objects, and with tracked objects the detector's hand flag hurts (+.025 \vs\ +.045).
Something-Else lists a clip's objects in the order of the caption's placeholders (categories match the caption order in 92.4\% of training videos), so the hand-last order also fixes each object slot's role.
At five seeds, randomising the object order with the hand last costs .027 (\twoSE\ .026; lower for every seed), and moving the hand first costs about as much again (.025, \twoSE\ .028); this split is post hoc.
The role order mainly helps \rel{inside} and \rel{above}; \INVZ barely changes across the three orders (.627--.640) while \INVS moves from .600 to .651, so a stable slot per role matters more for the relational channels than for the whole latent (descriptive).
Ordering tracked objects by the caption's nouns (Grounding DINO tiny on the placeholder strings, no boxes read) puts 70.0\% of object pairs in role order, against .381 for the detector's score order, short of the locked .75 bar; nothing was trained.
For planning with annotated boxes (5 relations, $K=8$, so not like-for-like with \cref{fig:use}), a probe on \INVS's relational channels plans at .340 top-5 with the objects in random order, against .166 for \PRED's whole latent (+.174, \twoSE\ .022, positive in every seed); caption order adds +.051 and raises \PRED by about as much (+.049, descriptive).
Given-box levels therefore include role information from the annotations, but the advantage that grounding gives does not depend on it, and since the format is identical across arms it changes the level of grounding, not the comparison.

\begin{table}[!htb]
\centering
\scriptsize
\setlength{\tabcolsep}{2.6pt}
\begin{tabular}{@{}l cccc@{}}
\toprule
Format & A$\to$A & A$\to$T & T$\to$A & T$\to$T \\
\midrule
Objects first, hand last, hand flag & {.103} & {.078} & .028 & .025 \\
Hand first, no flag (tracker format) & .063 & .051 & \best{.057} & \best{.052} \\
Objects first, hand last, no flag & \best{.116} & \best{.090} & {.052} & {.045} \\
Hand first, hand flag & .060 & .044 & \na & \na \\
Random object order, hand last, no flag & .089 & .068 & \na & \na \\
\bottomrule
\end{tabular}
\caption{\textbf{Grounding gain by input format}, training objects $\to$ test objects (A annotated, T tracked; 65 templates, 5 relations). The A$\to$A and A$\to$T cells of the three no-flag rows and the T$\to$A cell of the tracker-format row use 5 seeds, the others 3.}
\label{tab:format}
\end{table}

\paragraph{Clip quality, partial annotation and prompt.}
Training only on the half of the clips the tracker handles best (tracks present at both frames; no annotations used) grounds +.055, the other half +.033 (difference +.022, \twoSE\ .013, 5 seeds), a random half +.039, and all clips +.050.
Annotating 10\% or 25\% of the training videos adds nothing measurable in either format.
With ``object.'' alone as the prompt the detector finds 28\% of annotated hands, every arm's probe drops to about .72 (below the L4 probe bar, so the run is uninformative), and the gain on the pairs found is +.054.

\section{RH20T robot episodes}
\label{sec:rh20t}

\paragraph{Data.}
RoboInter's copy of RH20T, all 83 chunks (91.7\,GB), with its gripper and object boxes.
Windows of $k=30$ frames (3\,s at 10\,Hz) from all primary camera views, up to two pairs per episode and view; 98,395 training and 25,104 validation pairs from 9,181 and 2,403 episodes over 112 and 34 tasks.
The action is the 6-D end-effector displacement plus the change in gripper command, in the robot base frame, entering the predictor through a linear layer; the inverse head regresses it.

\paragraph{Box check.}
34 of 40 audited frames were correct (one object box on a cloth pattern; five boxes too small or occluded to judge), against a planned bar of 36.
The looser bar ($\ge 34/40$, at most two clearly wrong) was set after this audit; it and the rule that a failure would make the experiment inconclusive were fixed before any training.
Boxes from an off-the-shelf detector and tracker failed the original, stricter check (36 of 40 frames, no box on the background) before any training.
Grounding DINO base and SAM~2.1 large tracks (prompt ``robot gripper. object.''), run on a fixed sample of 1,999 validation pairs, passed the check in 5 of 40 audited frames: their gripper boxes covered the arm, or the gripper together with what it held.
Taking the gripper box from the robot's recorded pose and RH20T's camera calibration, and prompting the detector with each task's object nouns, fixed the gripper (fingertip boxes in about 34 of 40 frames) and raised manipulated-object coverage at IoU $\ge .5$ from 19\% to 31\%, but frame-sized or merged object boxes remained, and the check reached only 18 of 40.
The robot result therefore rests on the given boxes, and how much it depends on them is untested.

\paragraph{Grounding.}
+.081 (\INVSHUF $-$.009; probes .90--.95); the margin rule passes with a seeded random head (+.084, \twoSE\ .038).
Forecasts still depend on the action ($-$.095 when it is permuted) but reach only .518 AUROC on changing relations.
With one camera per robot setup, which keeps about a sixth of the pairs, the gain is +.046 at five seeds (seeds 3--4 added under \Crule{6}); the margin rule still passes, \INVS beats \INVSHUF by .094 (\twoSE\ .018), the probes read .81--.90, and \rel{left_of} still does not ground (\cref{tab:leftof}).

\paragraph{Why \rel{left_of} does not ground.}
A probe reads \rel{left_of} at .93 from \PRED's latent and .84--.86 from the entity latents, but no relational channel in any arm exceeds .58.
\cref{tab:leftof} lists the five explanations we tested.

\begin{table}[!htb]
\centering
\scriptsize
\setlength{\tabcolsep}{2.4pt}
\begin{tabularx}{\linewidth}{@{}X X l@{}}
\toprule
Explanation & Test & Result \\
\midrule
Captions name no direction & Real controls & \rel{left_of} $-$.031 \\
Cameras disagree on image left/right & One camera per setup, 5 seeds & \rel{left_of} $-$.013 \\
The action carries no left/right information & Mutual information per relation & second-most of five \\
The action is in the wrong frame & Controls mapped into each camera's image axes & no gain in information (ratio .90); not trained \\
Too few channels & $K=32$, on Something-Else & \rel{left_of} +.014 \\
\bottomrule
\end{tabularx}
\caption{\textbf{Explanations for \rel{left_of} not grounding.} The first four were tested on RH20T and the capacity test on Something-Else (40 templates, 13 relations), where with $K=32$ the extra channels go to relations that already ground.}
\label{tab:leftof}
\end{table}

\paragraph{Planning.}
Top-5 among 40 candidate controls: \INVS's relational channels .181, \PRED's .140, prior .125, model-free ridge oracle .237; a probe on any arm's whole latent .213--.229 (.229 for \PRED and \INVS, within .008 of the oracle); with the goal given as target channel values, \INVS\ .371 against \PRED\ .234.
The goals are mostly positional (\rel{near} holds in 98\% of pairs) and positions enter every arm's entity latents directly.
The relational channels hold about two-thirds of \INVS's planning information on RH20T (share .64; \PRED\ .17), against all of it on Something-Else.
We therefore claim relational planning for Something-Else only.

\section{Label-free objectives and label fractions}
\label{sec:labels}

\begin{figure}[!htb]
\centering
\includegraphics[width=\linewidth]{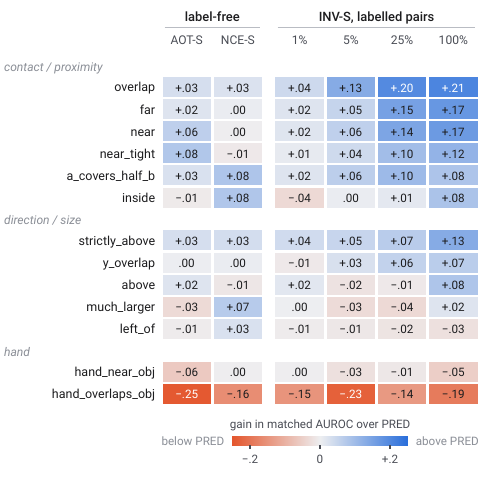}
\caption{\textbf{Per-relation gain over \PRED without action labels and with labels on part of the pairs} (65 templates, 5 seeds). Left: label-free losses on the channels, arrow of time (\AOT) and InfoNCE (\NCE). Right: \INVS with action labels on 1--100\% of the training pairs (locked design); the 100\% column is \INVS itself.}
\label{fig:s_labels}
\end{figure}

\paragraph{Label-free objectives.}
\emph{Arrow of time} (\AOT): each pair is presented in its true or reversed order at random, and a one-logit head with \INVS's inverse-head architecture classifies the order from the relational channels (binary cross-entropy).
\emph{InfoNCE} (\NCE): projections of the start-frame and goal-frame relational channels are compared by cosine similarity with temperature 0.1, and the other 63 clips in the batch are negatives.
Loss weights were set so that each loss matches the inverse loss at initialisation ($\lambda_{\text{aot}}=5.98$, $\lambda_{\text{nce}}=1.00$).
A held-out training-split check before launch gave arrow-of-time accuracy .688 and InfoNCE top-1 .899; on validation, .661 and .839.
Gains over \PRED are $-$.007 and +.010 at five seeds (the locked seed rule fired at three); both fall short of \INVS by .072 (\twoSE\ .005) and .056 (\twoSE\ .008).
Each gives a small gain on a few relations (arrow of time: \rel{near} +.061, \rel{near_tight} +.077; InfoNCE: \rel{inside} +.080, \rel{a_covers_half_b} +.081, \rel{much_larger} +.070); on contact and proximity as a group they gain +.035 and +.029, against +.138 for inverse dynamics (\cref{fig:s_labels}).

\paragraph{Label fractions.}
Labelled pools of 153, 701 and 3,490 pairs (1\%, 5\%, 25\%) are nested, sampled by video and stratified by template.
Each step draws 64 pairs from all pairs for the predictor and 64 from the labelled pool (with replacement) for the inverse loss, at weight 1, within the main setup's 8,720-step budget.
In the masked design, hashed in an addendum before any result, unlabelled pairs give the predictor a learned 64-d ``unknown action'' embedding, in both \INVS and a \PRED baseline at the same fraction.
\cref{tab:fractions} gives both designs.
The gain grows by .011 per doubling of labels (\twoSE\ .002), flat up to 5\%.
At 25\%, planning holds every rule (\PlChan\ .280 against the .174 prior; \PlS\ .354 against \PRED's whole-latent .242) and forecasting misses its bar narrowly (m2 +.046, bar .05); at 1\% every planning rule fails.
Contact and proximity ground first (\cref{fig:s_labels}, right): \rel{overlap}, \rel{near}, \rel{far}, \rel{near_tight} and \rel{a_covers_half_b} recover most of their full-label gain by 25\% (\rel{overlap} +.197 of +.207), whereas \rel{above}, \rel{inside} and \rel{strictly_above} need the full label set.
With the same 3,490 labelled pairs, \INVS reaches .641 when the other 75\% of the videos are present without labels and .643 when trained on those pairs alone, while \PRED rises from .548 to .595: the labelled pairs set \INVS's level, and the unlabelled pairs lift \PRED (a reading, not a tested rule).

\begin{table}[!htb]
\centering
\scriptsize
\setlength{\tabcolsep}{1.6pt}
\begin{tabular}{@{}l ccc c@{}}
\toprule
 & 1\% (153) & 5\% (701) & 25\% (3,490) & 100\% \\
\midrule
Gain, locked\up & $-$.002 $\pm$ .011 & {+.008} $\pm$ .013 & {+.047} $\pm$ .009 & +.066 $\pm$ .011 \\
Gain, masked\up & $-$.002 $\pm$ .010 & {+.009} $\pm$ .015 & {+.047} $\pm$ .010 & \na \\
\INVS, masked & .596 & .607 & .645 & \na \\
\PRED, masked & .597 & .598 & .598 & \na \\
Inverse top-1, locked\up & .083 & {.122} & {.189} & .318 \\
\bottomrule
\end{tabular}
\caption{\textbf{Label fractions} (65 templates, 13 relations, 5 seeds; gain $\pm$\twoSE). No fraction below 100\% passes \Crule{1}; 25\% misses it by .003 and holds \Crule{2}--\Crule{4}. The masked design hides the unlabelled pairs' actions from the predictor too.}
\label{tab:fractions}
\end{table}

\section{Planning and the distributed code}
\label{sec:planning}

\cref{tab:planning} gives the planners for every arm on the main setup.

\begin{table}[!htb]
\centering
\scriptsize
\setlength{\tabcolsep}{2.8pt}
\begin{tabular}{@{}l cccccc@{}}
\toprule
Arm & \PlDist\up & \PlChan\up & \PlInv\up & \PlZ\up & \PlS\up & \PlR\up \\
\midrule
\PRED & .046 & .067 & \na & .244 & .168 & .214 \\
\INVSHUF & {.118} & .136 & \na & .237 & .184 & .213 \\
\INVR & .075 & .113 & \na & .311 & .209 & .301 \\
\INVZ & .101 & {.297} & \best{.482} & {.391} & {.380} & \best{.329} \\
\INVS & \best{.146} & \best{.334} & {.374} & \best{.416} & \best{.425} & {.316} \\
\midrule
\textcolor{ceilinggray}{\SUP} & \na & \textcolor{ceilinggray}{.394} & \na & \na & \textcolor{ceilinggray}{.393} & \na \\
\bottomrule
\end{tabular}
\caption{\textbf{Planning from relational goals}, top-5 among 65 templates, 13 relations, 3 seeds, readouts fitted on all labelled training pairs. \PlDist: squared distance of the matched channels; \PlChan: one calibrated channel per relation; \PlInv: the arm's inverse head; \PlZ, \PlS, \PlR: probes on the forecast $z$, $s$ or $r$. Prior .174, chance .077. \up: higher is better; bold: best arm without relation labels}
\label{tab:planning}
\end{table}

\paragraph{Readout labels.}
The readouts need few labelled pairs (\cref{fig:use}c): with 1\% of the training pairs (about 140), \INVS's channel probe plans at .415 and its one-channel readout at .332, while \PRED's whole-latent probe needs all pairs to reach .244 (.179 at 1\%).

\begin{figure}[!htb]
\centering
\includegraphics[width=\linewidth]{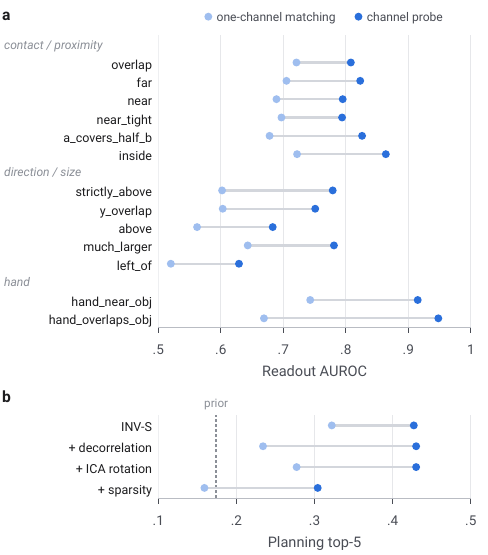}
\caption{\textbf{The relational code is spread over the relational channels}. (a) \INVS's readout AUROC for each relation from the one channel matched to it (light) and from a probe on all 16 relational channels (dark). (b) Planning top-5 with the one-channel readout \PlChan (light) and the channel probe \PlS (dark), for \INVS and three label-free attempts to align relations with channels (the ICA rotation is applied after training); dashed: the frequency prior. \INVS is retrained here on the variants' code path (5 seeds), so its values differ slightly from \cref{tab:planning}.}
\label{fig:s_code}
\end{figure}

\paragraph{The code is distributed.}
In the grounded model the relational information sits in the 16 relational channels, but each relation is spread over several of them (\cref{fig:s_code}a): the one-channel readout \PlChan plans at .334 and the 16-channel probe \PlS at .425.
Three label-free attempts to give each relation its own channel failed (\cref{fig:s_code}b).
A \emph{decorrelation penalty} (the mean squared off-diagonal entry of the channel correlation matrix, with its weight chosen without relation labels, on held-out inverse accuracy) decorrelates the channels (mean $|$off-diagonal correlation$|$ .24 to .014) but makes the one-channel readout worse (readout AUROC $-$.037; \PlChan\ .322 to .234) and drops the grounding gain to +.028, below the bar; nothing is lost from the code as a whole (\PlS\ .430 \vs\ .427; \PlInv\ .368 to .467).
Decorrelation is invariant to rotations of the code, so it can whiten it but not align relations with channels; the penalty alone on \PRED lowers its matched AUROC by .038.
A post-hoc \emph{ICA rotation}, which is rotation-sensitive, also makes the one-channel readout worse (readout AUROC $-$.024, \twoSE\ .015; \PlChan\ .322 to .277, within noise), and a \emph{sparsity penalty} cuts active channels per pair from 8 to 2.4 but destroys grounding (+.005), planning (\PlChan\ .159, \PlS\ .304) and the action effect.
The inverse-trained channel axes remain the best one-channel basis we found, and matched AUROC is a lower bound on what the channels hold.

\paragraph{Rollout.}
Rolling the predictor forward within a clip in steps of about 1\,s (thirteen relations), the action effect grows from +.021 after one step to +.051 after two and +.057 after three, and the forecasts beat copying; trained on these shorter steps, grounding falls below the bar (+.037, with \Crule{2}--\Crule{4} holding), and planning from relational goals does not beat the prior.

\section{Frozen V-JEPA~2 encoder}
\label{sec:vjepa}

\paragraph{Setup.}
Frozen V-JEPA~2 ViT-L, self-supervised weights only; entity features are RoIAlign-pooled from the full-frame token map (16$\times$16 grid) at each box, not taken from box crops; otherwise the main setup (65 templates, 13 relations).
With video context, 16 past frames at 12\,fps end at each frame and features come from the last temporal token slice, the one containing that frame, so the boxes stay aligned.
V-JEPA~2's pretraining mixture includes 168K Something-Something V2 videos, used without labels~\cite{assran2025vjepa2}, about the size of SSv2's training split (168,913 videos~\cite{materzynska2020somethingelse}).
Something-Else's compositional split assigns videos to training and validation by verb--object combination rather than by SSv2's split~\cite{materzynska2020somethingelse}, so 13,057 of the 14,975 validation videos used here (87\%) are SSv2 training videos, which the encoder saw during pretraining; the other 1,918 are SSv2 validation videos.
Every arm shares the frozen encoder, so the arms are compared like for like.
To test whether this exposure inflates the scores, we re-scored the 25 still-frame checkpoints (five arms, five seeds) on the 1,918 unseen videos and, as a size-matched reference, on five random 1,918-video samples of the 13,057 seen ones.
\INVS's margin gain over \PRED is +.042 $\pm$ .010 on unseen videos and +.045 $\pm$ .016 on the matched samples (seed-paired difference $-.002 \pm .007$; all $\pm$\twoSE), so the gain is unchanged.
Only \INVZ meets the locked rule for inflation by exposure (a margin lower on unseen than on matched seen videos by more than \twoSE): its margin falls from .088 on the matched samples to .082 on unseen videos ($-.005 \pm .003$), and the ranking of the arms does not change.
Seen and unseen videos also come from different SSv2 splits, so these differences combine exposure and split.
The V-JEPA~2 ceiling in \cref{sec:ceiling} was not re-scored and may be optimistic.

\begin{table}[!htb]
\centering
\scriptsize
\setlength{\tabcolsep}{1.8pt}
\begin{tabular}{@{}l cc c l c@{}}
\toprule
Encoder input & \PRED & \INVS\up & Gain\up & \PlS\up \\
\midrule
ResNet-18, still frames & .593 & {.658} & {+.065} & {.425} (.244) \\
V-JEPA~2, still frames & .599 & {.648} & {+.050} & .398 (.257) \\
\quad still, 5 seeds & .601 & .643 & +.042  & .390 (.252) \\
\quad 16 past frames & .625 & .641 & +.016  & {.399} (.308) \\
\quad video start, still goal & .619 & .640 & +.021 & .398 (.294) \\
\bottomrule
\end{tabular}
\caption{\textbf{Encoder} (65 templates, 13 relations; matched AUROC). Planning: \INVS's channel probe, top-5, with \PRED's whole-latent probe in brackets. The first two rows use 3 seeds, the others 5. \up: higher is better}
\label{tab:vjepa}
\end{table}

\paragraph{Results} (\cref{tab:vjepa}).
Still frames: gain +.0496 at three seeds, short of \Crule{1} by .0004, and +.042 $\pm$ .015 (\twoSE) at five; contact and proximity ground as with ResNet-18 (\rel{far} +.20, \rel{overlap} +.18, \rel{near} +.14), while the vertical relations lose their gain.
Video context: +.016 at five seeds, while the inverse head's action accuracy rises beyond \twoSE (\INVS\ .496 against .458), the outcome the locked reading attributed to the action becoming readable from motion rather than from relational change.
A logistic action probe on entity features names the action from the goal frame alone at .566 with video context against .479 with still frames (+.087, \twoSE\ .008); this was measured in the same locked run but outside its rules, which by a specification error tested the start frame, and is descriptive.
A video start frame with a still goal frame gives +.021; inverse accuracy returns to the still-frame level, but grounding does not.
At three seeds, seeding the random head turns \Crule{2} from fail to pass with video context (margin gain .031, \twoSE\ .016) and with a video start frame (.028, \twoSE\ .014); at five seeds \Crule{2} holds either way, but \Crule{1} fails in both, so the verdicts stand.
With video context \INVS stays at .640--.641 while \PRED rises to .619--.625, so there the gain shrinks mainly because the baseline improves; with still frames \PRED barely changes from ResNet-18 (.599 \vs\ .593) and \INVS is lower (.648 \vs\ .658), so that miss has no single cause. This is a reading across experiments rather than one pre-registered test.

\begin{figure}[t]
\centering
\includegraphics[width=\linewidth]{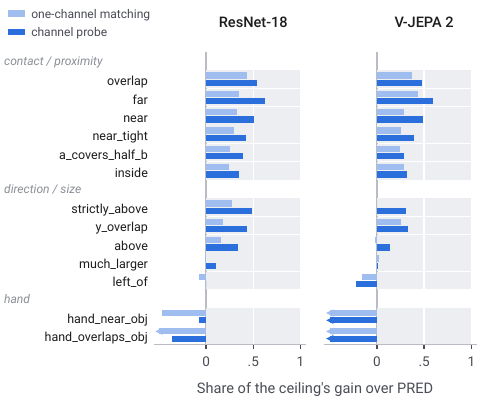}
\caption{\textbf{Share of the supervised ceiling} that inverse dynamics recovers, $(\INVS-\PRED)/(\SUP-\PRED)$ in matched AUROC (light) or channel-probe AUROC (dark): 0 is \PRED and 1 the label-trained ceiling (grey track). Light: one channel matched to each relation; dark: a probe on all 16 relational channels. Bars cut at $-$.5 continue to between $-$.61 and $-$1.97 (hand relations, whose ceiling gain over \PRED is small). Ceiling 3 seeds, other arms 5.}
\label{fig:s_ceiling}
\end{figure}

\section{Supervised ceiling}
\label{sec:ceiling}

\SUP is the same model with relational channel $\rho$ trained with a binary cross-entropy on relation $\rho$ (13 of 16 channels), on ResNet-18 and V-JEPA~2, 3 seeds; a second arm removes the box coordinates from the entity input.
Its locked precondition, a ceiling of at least .85 on ResNet-18, holds (.998; V-JEPA~2 .997): the relations are a near-exact function of the box coordinates in the input.
Share of the ceiling's gain over \PRED recovered by inverse dynamics ($\pm$\twoSE; ceiling 3 seeds, baselines 5): one-channel matching .160 $\pm$ .032 (ResNet-18) and .123 $\pm$ .067 (V-JEPA~2); channel probe .371 $\pm$ .011 and .252 $\pm$ .067.
By group (matching / probe): contact and proximity .32 / .48 (ResNet-18) and .31 / .44 (V-JEPA~2); direction and size .11 / .28 and .03 / .12; little or none for \rel{left_of}, \rel{much_larger} (at most .10) and the hand relations; \cref{fig:s_ceiling} gives every relation.
Label-trained channels show two to three times the action effect when forecasting relational change (.223 \vs\ .103 with ResNet-18; .227 \vs\ .080 with V-JEPA~2) and plan better with the one-channel readout (\PlChan\ .394 \vs\ .331; .406 \vs\ .264), but not through a channel probe (\PlS\ .393 \vs\ .424; .398 \vs\ .390).
Without box coordinates, label-trained channels on V-JEPA~2 features still read every relation that inverse dynamics missed (.988--.998), so the miss is the objective's, not the encoder's; this measures position available through positional embeddings, not relational understanding.
On ResNet-18 without boxes, \rel{left_of} reaches .685, \rel{above} .765 and \rel{strictly_above} .848.

\section{Pre-registration record}
\label{sec:registry}


\paragraph{Decision rules.}
Before the first real-video result we fixed six rules that decide whether a setting passes; the main text reports the numbers behind them rather than the verdicts.
\Crule{1}: gain $\ge .05$.
\Crule{2}: \INVS's margin is positive and exceeds \PRED's by more than \twoSE.
\Crule{3}: \INVSHUF does not reproduce the gain, and \INVS beats it by more than \twoSE.
\Crule{4}: the probe reads the relations at AUROC $\ge .75$ in every arm; otherwise the run is uninformative, not negative.
\Crule{5}: \Crule{1}--\Crule{3} hold on the compositional validation split.
\Crule{6}: if the gain is within three seed SDs (pooled over \PRED and \INVS) of zero, seeds 3--4 are added and the five-seed verdict is final.
A setting passes only if all rules hold.
Each later experiment fixed its own rules, usually these, before its first result; experiments with tracked objects use the analogous rules L1--L4.

\paragraph{Seeded random heads.}
The evaluator originally drew the random head behind \Crule{2}'s margin (L2 for tracked objects) without a fixed seed, so a checkpoint's margin varied between evaluations while its matched AUROC reproduced exactly; across nine further seeds, a single draw moves a checkpoint's margin by up to about .13.
We therefore re-scored all 812 real-video checkpoints trained up to that point with the random head seeded at 0, using one evaluator.
Every matched AUROC reproduces (largest difference $2.7{\times}10^{-5}$) and no overall verdict changes.
Seeding roughly halves the \twoSE of margin-gain comparisons.
This turns \Crule{2} from fail to pass in the three-seed comparisons of the two V-JEPA~2 runs with video input (at five seeds it held either way, and \Crule{1} fails in both) and from pass to fail in the three-seed comparison with action labels on 5\% of the pairs (already negative on \Crule{1}), turns one descriptive margin comparison with tracked objects from fail to pass, and changes the verdict on \INVR in two five-relation settings from no grounding to partial grounding, because its small margin gain over \PRED (about +.02) now clears \twoSE.
All margins in this paper use the seeded random head unless stated otherwise; two tracker evaluations could not be re-scored because their track alignments were not stored, and later experiments used the seeded evaluator from the start.
Each margin rests on one random head (seed 0); averaged over nine further seeds, every experiment's mean margin lies within .025 of its seed-0 value.

\end{document}